%% file: amazon_arxiv_v1.tex
\documentclass[twoside]{article}

\usepackage{amazon_paper}

\renewcommand{\bibname}{References}
\renewcommand{\bibsection}{\subsubsection*{\bibname}}

\usepackage{amssymb}
\usepackage{xcolor}
\usepackage{booktabs}
\usepackage[]{graphicx}
\usepackage{caption}
\usepackage{subcaption}
\usepackage{wrapfig}
\definecolor{amazondarkblue}{HTML}{232F3E}

\usepackage{multirow}

\usepackage[table]{xcolor}
\usepackage{multirow}
\usepackage{array}
\definecolor{baselinegray}{gray}{0.92}
\definecolor{overallbg}{RGB}{238,242,255}
\newcolumntype{O}{>{\columncolor{baselinegray}}c}

\usepackage{pifont}
\usepackage{makecell} 
\usepackage[dvipsnames]{xcolor} 

\usepackage{listings}
\usepackage{xcolor}
\usepackage{algorithm}
\usepackage{algpseudocode}
\usepackage{threeparttable}
\usepackage{enumitem}

\lstdefinelanguage{json}{
    basicstyle=\ttfamily\small,
    comment=[l]{//},
    morestring=[b]",
    stringstyle=\color{blue},
    literate=
     *{0}{{{\color{black}0}}}{1}
      {1}{{{\color{black}1}}}{1}
      {2}{{{\color{black}2}}}{1}
      {3}{{{\color{black}3}}}{1}
      {4}{{{\color{black}4}}}{1}
      {5}{{{\color{black}5}}}{1}
      {6}{{{\color{black}6}}}{1}
      {7}{{{\color{black}7}}}{1}
      {8}{{{\color{black}8}}}{1}
      {9}{{{\color{black}9}}}{1}
}

\usepackage[most]{tcolorbox}
\usepackage{courier} 

\newtcolorbox{promptbox}[2][]{%
  enhanced,
  breakable,
  colback=gray!2,
  colframe=gray!30,
  boxrule=0.4pt,
  arc=6pt,
  outer arc=2pt,
  fontupper=\ttfamily\small,
  title={#2},
  coltitle=white,
  colbacktitle=gray!80!black,
  attach boxed title to top left={yshift=-1.5mm, xshift=2mm},
  boxed title style={
    sharp corners,
    boxrule=0pt,
    colback=gray!80!black,
    colupper=white,
    fontupper=\bfseries\footnotesize,
  },
  #1
}

\begin{document}

\title{%
\raisebox{-2.5em}{%
  \parbox[t]{1.0in}{\includegraphics[width=0.9in]{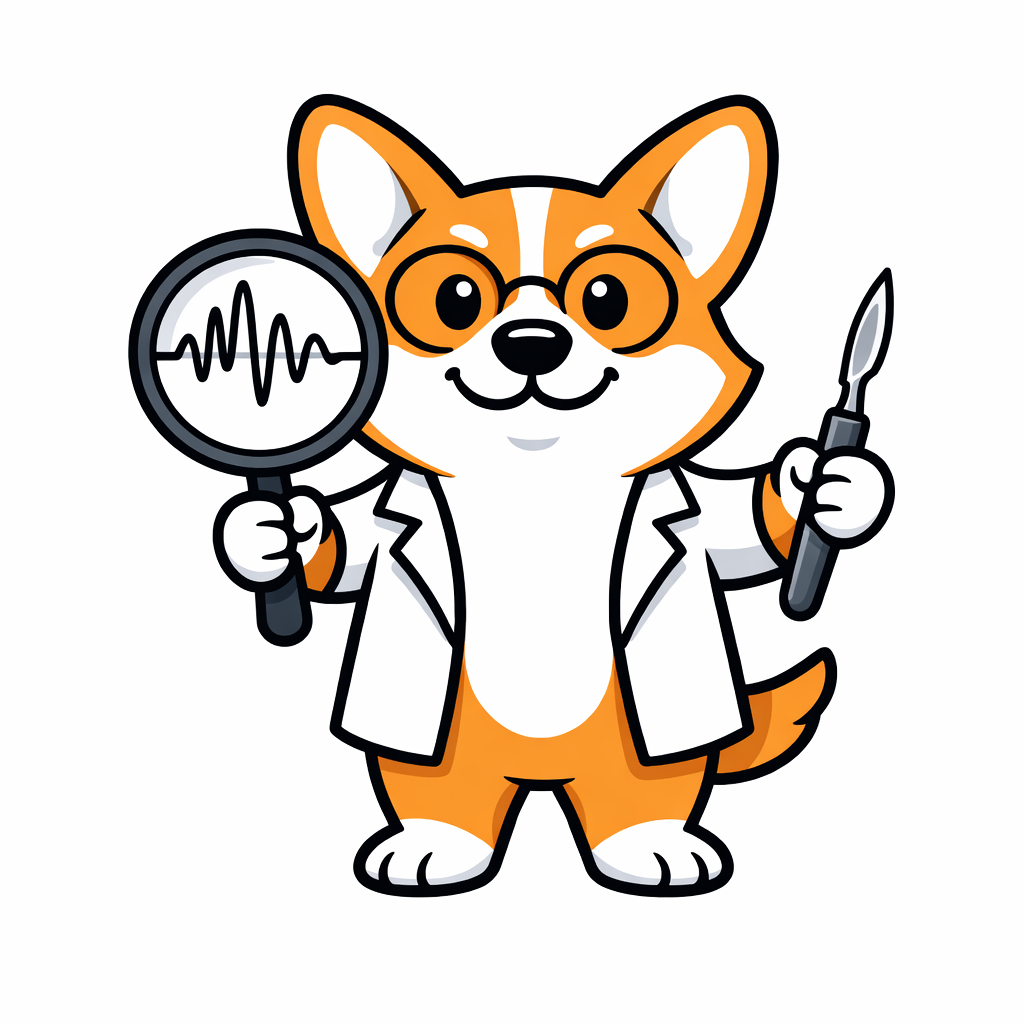}}%
}%
\hspace{0.1in}%
\parbox[t]{0.7\textwidth}{\centering SD-MAR: Multi-image Analytical Reasoning via Synthetic Data and Reinforcement Learning}%
}

\author{%
\small{Shiyu Yuan$^{1}$\thanks{Work performed during an internship at Amazon Web Services (AWS) Bedrock.}, 
Sourav Sanjukta Bhabesh$^{2}$,
Zhe Wang$^{2}$, \\
Dmitriy Bespalov$^{2}$,
Wesley Rose$^{3}$,
Huzefa Rangwala$^{3}$\thanks{Work done while at Amazon Web Services (AWS).}}%
\\[1em]
{\fontsize{10pt}{11pt}\selectfont
$^{1}$Stevens Institute of Technology \quad
$^{2}$AGI Foundations for AWS \quad
$^{3}$Amazon Web Services (AWS), Amazon}
}

\maketitle
\vspace{-1.5em}
\definecolor{amazonorange}{HTML}{FF9900}
\renewenvironment{abstract}{%
  \begin{tcolorbox}[colback=amazonorange!8, colframe=black, boxrule=0.5pt, arc=3pt, left=10pt, right=10pt, top=8pt, bottom=8pt]
  \centerline{\textbf{Abstract}}\vspace{0.5em}
}{%
  \end{tcolorbox}
}

\begin{abstract}
Vision Language Models (VLMs) demonstrate strong perceptual abilities but remain limited in tasks that require analytical reasoning across multiple visual states, such as multi-image comparison, change detection, and multi-step analytical inference. These capabilities are critical for real-world multimodal applications where reasoning must be grounded in systematic differences between visual contexts. However, existing benchmarks rarely require both explicit visual comparison and analytical reasoning, leaving this capability not explored sufficiently. To address this gap, we introduce SD-MAR (Synthetic Data for Multi-image Analytical Reasoning), a framework designed to  train and evaluate models on visual comparison and analytical reasoning. SD-MAR constructs paired visual scenarios through controlled perturbations and generates reasoning tasks across two domains: semantic change attribution and quantitative comparison reasoning. To train VLMs on SD-MAR, we employ GRPO-lite with Backward Discounted Allocation (BDA), a reinforcement learning configuration purpose-built for multi-image analytical reasoning. GRPO-lite removes KL regularization to relax the constraint that keeps the policy close to the pretrained distribution, and applies backward-discounted reward allocation to assign greater credit to the decisive later reasoning steps where analytical conclusions are drawn. Experiments on Qwen2.5-VL-7B and InternVL3-8B show that GRPO-lite fine-tuning on SD-MAR yields up to 36.95\% in-domain accuracy improvement, with Qwen2.5-VL-7B outperforming GPT-4.1 on the SD-MAR test benchmark. Crucially, out-of-domain generalization is preserved or improved: models maintain performance within $\pm$1\% on MME, MMMU-Pro, and MathVista, while gaining up to 4\% on MMBench. LLM-as-judge evaluation further confirms improved reasoning quality, with consistent gains in logical coherence and explanation clarity across both models.
\end{abstract}

\input{sec/1_intro_colm_sourav_v3}

\input{sec/4_datasets_arxiv_v0}
\input{sec/5_exp_colm_arxiv_v0}
\input{sec/7_conclusion_colm_sourav_v3}

\clearpage
\bibliography{colm2026_conference}
\bibliographystyle{colm2026_conference}

\appendix
\clearpage
\section{Appendix}
\input{sec/X_suppl_amazon_arxiv_v1}

\end{document}

%% file: sec/1_intro_colm_sourav_v3.tex
\section{Introduction}
\label{sec:intro}
Vision Language Models (VLMs) have achieved strong performance on perception-oriented multimodal understanding tasks such as image captioning, visual question answering, and instruction following \citep{radford2021learning, li2023blip, liu2023visual}. Despite these advances, current models remain limited in tasks that require analytical reasoning across multiple visual states, including multi-image comparison, change detection, and multi-step analytical inference \citep{huang2024vbench, fu2023mme, li2023seed, lu2023mathvista}. 

In real-world applications, reasoning must be grounded in systematic differences between visual contexts rather than isolated image understanding. Examples include detecting temporal changes in medical imaging, monitoring environmental variations in satellite imagery, and reasoning about evolving scenes in robotics. These tasks require models to interpret multiple visual states, identify differences between them, and perform inference grounded in those differences. We refer to this capability as \emph{multi-image analytical reasoning} (MAR), defined as the ability to compare visual states, detect changes, and conduct multi-step inference based on those changes.

A central challenge in studying MAR arises from limitations in existing datasets and evaluation benchmarks. Many multi-image multimodal datasets are created with weakly controlled perturbation and primarily designed for perception-level understanding or single-step reasoning. As a result, they rarely isolate tasks that require systematic comparison across visual states or analytical reasoning over controlled perturbations. Several benchmarks explore related capabilities, including multi-image reasoning (NLVR2 \citep{suhr2019nlvr2}), compositional visual reasoning (CLEVR \citep{johnson2017clevr}, GQA \citep{hudson2019gqa}), and robustness under perturbations (ImageNet-C \citep{hendrycks2019benchmarking}). More recent multimodal evaluation suites such as MMBench \citep{liu2024mmbench}, MM-Vet \citep{yu2023mm}, MathVista \citep{lu2023mathvista}, V-Bench \citep{huang2024vbench}, SEED-Bench-2 \citep{li2024seed}, and BLINK \citep{fu2024blink} expand coverage across reasoning, robustness, and instruction following. However, these benchmarks primarily evaluate individual aspects of multimodal reasoning and do not explicitly integrate controlled visual transformations with multi-image comparison and multi-step inference within a unified evaluation setting.

To address this gap, we introduce \textbf{SD-MAR}, a scalable synthetic data generation framework designed to study multi-image  analytical reasoning under controlled visual perturbations. SD-MAR constructs paired visual scenarios through explicit attribute-level transformations and generates reasoning tasks that require models to attribute semantic changes and perform quantitative comparisons across visual states. By controlling perturbations during data generation, SD-MAR enables scalable construction of diverse reasoning instances while maintaining verifiable ground-truth supervision. This design allows the dataset to isolate analytical reasoning from confounding correlations commonly present in natural-image datasets.

We validate SD-MAR at two complementary levels. First, we evaluate the validity of the dataset through human annotation and analysis of the synthetic data generation (SDG) pipeline, ensuring that generated tasks accurately reflect the intended analytical reasoning. Second, we leverage SD-MAR multi-image perturbations to derive verifiable supervision for analytical multi-image reasoning, and fine-tune open-source VLMs using reinforcement learning with verifiable rewards (RLVR). We further examine whether such targeted supervision improves reasoning beyond superficial pattern matching through LLM-as-judge. 

Building on this framework, our work makes three main contributions. First, we formalize \emph{multi-image analytical reasoning} (MAR) as a capability that requires comparison across visual states, explicit change detection, and multi-step inference grounded in attribute perturbations. Second, we introduce SD-MAR, a synthetic data generation framework that instantiates this reasoning setting through controllable visual transformations and verifiable ground-truth supervision. Third, we adopt a purpose-built GRPO configuration with Backward Discounted Allocation (BDA), referred to as GRPO-lite, to better align the training procedure with the MAR setting.

To validate this formulation, we design targeted evaluation tasks that isolate perceptual comparison and analytical reasoning, and evaluate a diverse set of close- and open-source VLMs to quantify their performance and identify limitations in multi-image analytical reasoning. Through RLVR fine-tuning of open-source VLMs, we demonstrate that SD-MAR paired with GRPO-lite provides an effective training signal that substantially improves multi-image analytical reasoning performance while preserving general multimodal capability.

\subsection{Related Work}
\label{sec:related}
\paragraph{Single-image vision-language reasoning benchmarks.}
A number of benchmarks have been proposed to evaluate reasoning capabilities in VLMs under single-image settings. Datasets such as MMMU~\citep{yue2024mmmu}, MMBench~\citep{liu2024mmbench}, and MM-Vet~\citep{yu2023mm} evaluate multimodal reasoning, instruction following, and knowledge grounding across diverse tasks. MathVista~\citep{lu2023mathvista} emphasizes mathematical reasoning grounded in visual inputs, while V-Bench~\citep{huang2024vbench} and BLINK~\citep{fu2024blink} study fine-grained capability decomposition and robustness under visual perturbations. However, these benchmarks predominantly focus on reasoning over a single image and do not explicitly require comparison across multiple visual states.

\paragraph{Multi-image reasoning and change understanding.}
\label{sec:realted_multi_image}
Prior work has explored reasoning over differences between images, particularly in change detection and change captioning tasks. Models such as DUDA~\citep{park2019robust} generate textual descriptions of scene changes by attending to regions across image pairs. Synthetic datasets such as CLEVR-Change~\citep{qiu2021describing} introduce controlled object-level modifications to study visual change localization and description. More recent benchmarks, such as SEED-Bench-2~\citep{li2024seed}, incorporate tasks that require understanding relationships across multiple images. However, these datasets primarily focus on detecting or describing visual differences and do not explicitly encourage reasoning processes to be organized in an analytical progression grounded in those differences.

%% file: sec/4_datasets_arxiv_v0.tex
\section{SD-MAR framework}
\label{sec:datasets}
To address the limitations identified above, we propose the SD-MAR framework for studying the multi-image analytical reasoning (MAR) setting defined in Sec.~\ref{sec:mar}. The framework comprises two main modules: (a) a Synthetic Data Generation module (SD-MAR SDG) in Sec.~\ref{sec:sd_mar_sdg} and (b) a Reinforcement Learning with Verifiable Rewards module (SD-MAR RLVR) in Sec.~\ref{sec:methods}.

\subsection{Multi-image Analytical Reasoning (MAR)}
\label{sec:mar}
We define \emph{multi-image analytical reasoning} (MAR) as reasoning over differences between multiple visual states. Given a set of visual inputs $\mathcal{V}=\{v_1, v_2, \dots, v_n\}$ representing different states of a scene and a query $q$ that requires reasoning about the differences across these states, the goal is to infer an answer by comparing the visual inputs and analyzing the changes relevant to the query.

Let $z_i = g(v_i)$ denote the semantic representation of visual state $v_i$, and let differences between states be defined as $\Delta_{ij} = h(z_i, z_j)$, where $\Delta_{ij}$ captures attribute-level changes or relationships between visual states. The final reasoning output is obtained by performing inference over these visual differences, $a = f(\{\Delta_{ij}\}, q)$. MAR therefore requires models to interpret visual states, identify differences between them, and perform multi-step inference grounded in those differences.

\subsection{SD-MAR SDG: Module for Multi-image Analytical Reasoning Synthetic Data Generation}
\label{sec:sd_mar_sdg}
The SD-MAR SDG pipeline generates high-quality, analytical reasoning data for training and evaluating VLMs under controlled visual perturbations. Within this framework, we define two task categories.
(1) \textbf{SD-MAR-TSE} produces semantic reasoning tasks spanning \textbf{T}emporal, \textbf{S}patial, and \textbf{E}motional transformations (TSE). (2) \textbf{SD-MAR-Math} constructs quantitative, comparison-based reasoning tasks across three difficulty levels.



\paragraph{SD-MAR-TSE dataset} 
In the semantic domain, MAR manifests as the ability to interpret each image independently, identify differences between paired images, and infer the underlying factors responsible for those differences. 

\begin{figure}
    \centering
    \includegraphics[width=0.99\linewidth]{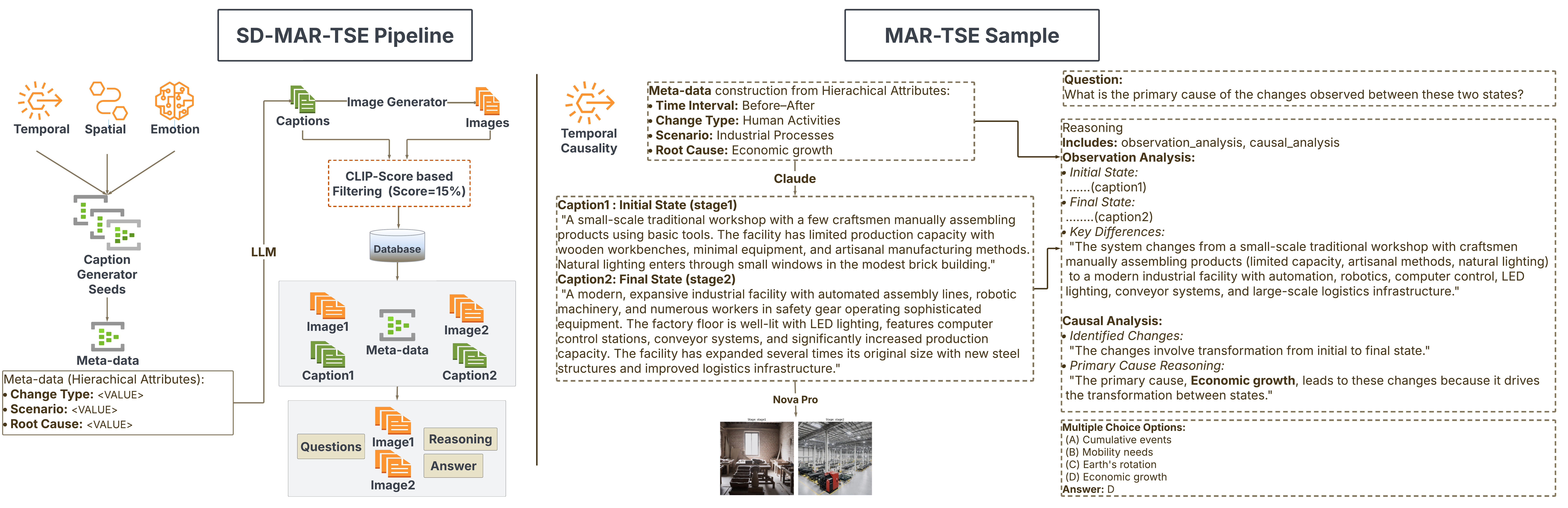}
    \caption{SD-MAR-TSE}
    \label{fig:sd_mar_tse}
\end{figure}

To construct high-quality semantic SD-MAR-TSE data, the SD-MAR data generation module employs a perturbation-controlled generation pipeline. It begins with a \emph{hierarchical caption seed generator} (see Appendix ~\ref{app:generator_seeds}) encoding domain-specific change relationships. Each sample is created by: a) selecting hierarchical attributes (e.g., change type, affected aspect, root factor); b) generating aligned captions and metadata using Claude Sonnet 3.7; and c) rendering paired images via an image generator (Amazon Nova-Pro). By modifying only targeted attributes while preserving others, the pipeline ensures fine-grained semantic control over the generated visual states. The structured metadata is injected into the language model to preserve cross-modal consistency. As illustrated in Fig.~\ref{fig:sd_mar_tse} and Alg.~\ref{alg:MAR_tse}, the pipeline constructs diverse, interpretable image pairs along with reasoning queries. We present controlled MAR temporal, spatial, and emotional samples, along with their grounded reasoning traces, in Appendix~\ref{app:temporal_samples}, \ref{app:spatial_samples}, and \ref{app:emotion_samples}, respectively.

\vspace{-3ex}
\begin{algorithm}[t]
\footnotesize
\caption{SD-MAR-TSE Generation}
\label{alg:MAR_tse}
\begin{algorithmic}[1]
\Require Reasoning type $r \in \{\text{Temporal},\ \text{Spatial},\ \text{Emotion}\}$; 
CLIP threshold $\tau$; LLM (Claude); Vision (Nova-Pro)
\State $\text{meta-data} \gets \Call{Sample Attributes}{r}$ \hfill \textcolor{gray}{// domain prompt: change type, change aspects, root cause}
\State $(I_1, I_2) \gets \Call{Nova-ProRender}{c_1, c_2}$ 
\State $\text{clip}_1 \gets \Call{CLIPScore}{I_1, c_1}$; $\text{clip}_2 \gets \Call{CLIPScore}{I_2, c_2}$
\If{$\min(\text{clip}_1, \text{clip}_2) < \tau$}
  \State \Return \Call{Reject And Resample}{} 
\EndIf
\State $y \gets \Call{ExtractRootFactor}{\Delta, r}$ 
\State $q \gets \Call{ConstructAnalyticalQueries}{(I_1,I_2), (c_1,c_2), \Delta}$ 
\State $\mathcal{O} \gets \Call{MakeSemanticDistractors}{y, r}$ 
\State $\text{meta} \gets \{\text{type}{=}r,\ \text{diffs}{=}\Delta,\ \text{captions}{=}(c_1,c_2),\ \text{clip}{=}(\text{clip}_1,\text{clip}_2)\}$
\State \Return $s{=}\{\text{inputs: }(I_1,I_2),\ \text{question: }q,\ \text{choices: }\mathcal{O},\ \text{answer: }y,\ \text{metadata: meta}\}$
\end{algorithmic}
\end{algorithm}

\begin{figure}
    \centering
    \includegraphics[width=0.99\linewidth]{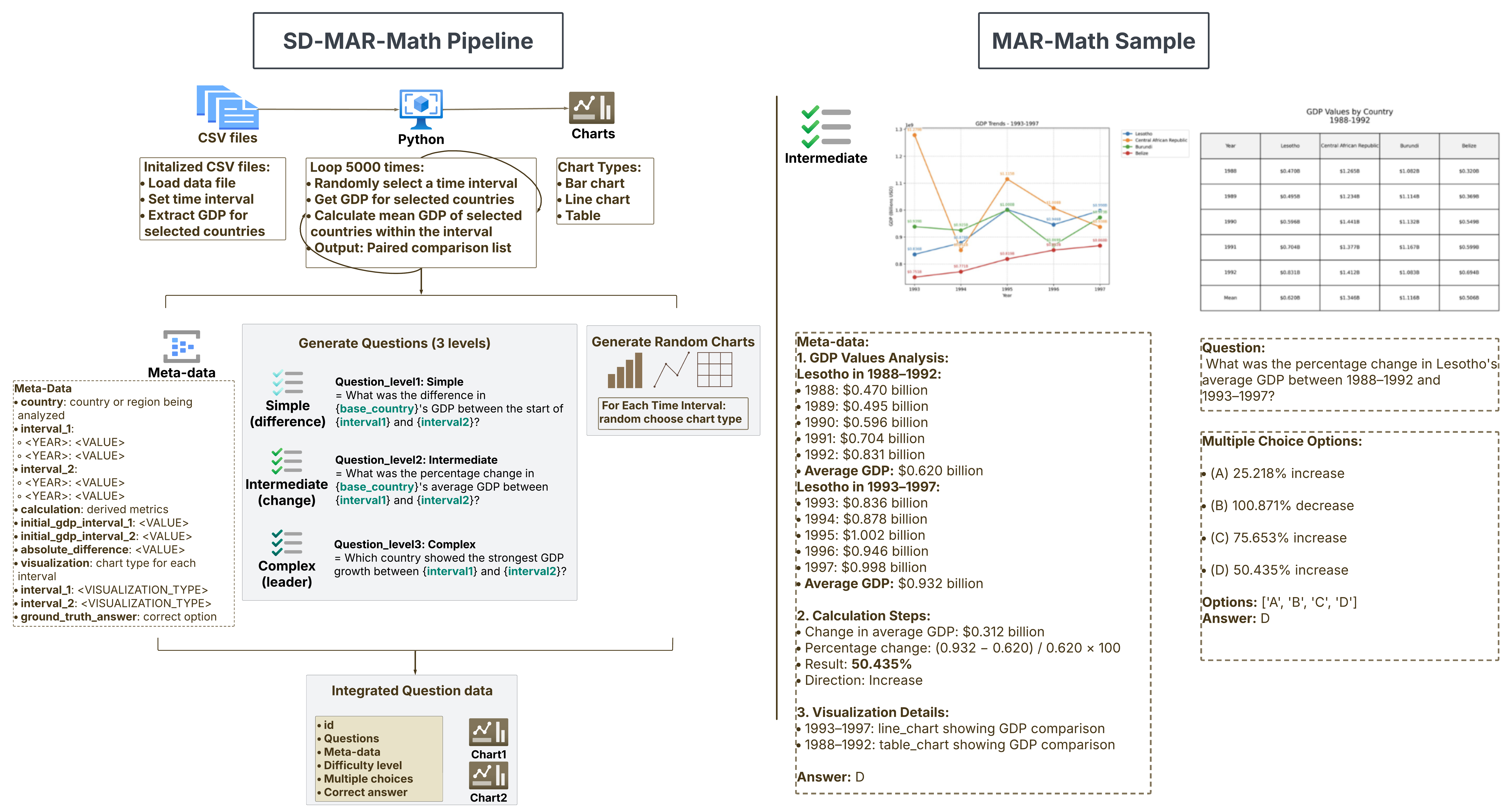}
    \caption{SD-MAR-Math}
    \label{fig:sd_mar_math}
\end{figure}

\paragraph{SD-MAR-Math datasets} 
We next construct the SD-MAR-Math dataset, which focuses on quantitative comparison-based analytical reasoning. We source real-world GDP time series data\footnote{World Bank: GDP time series, \url{https://wits.worldbank.org/CountryProfile/en/country/by-country/startyear/ltst/endyear/ltst/indicator/NY-GDP-MKTP-CD}} and generate paired visual comparison samples using disjoint temporal intervals. The paired-image setting introduces controlled temporal variation by comparing two disjoint time intervals. Charts and tables are rendered using \texttt{matplotlib} in financial-report styles (e.g., bar charts, line charts, tables). Fig.~\ref{fig:sd_mar_math} illustrates both the generation pipeline and a representative MAR-Math sample.

SD-MAR-Math tasks are stratified into three difficulty levels based on the number of reasoning hops required. 1) \emph{Level 1 (Simple)} computes directional GDP differences for a single country across intervals, 2) \emph{Level 2 (Intermediate)} first aggregates GDP values for a given country within each interval (e.g., by computing the average GDP across years), and then computes the percentage change between the intervals, 3) \emph{Level 3 (Complex)} identifies the country with the largest growth rate across intervals among multiple countries. The generation algorithm is detailed in Alg.~\ref{alg:MAR_math}. In Appendix~\ref{app:math_samples_trace}, we present more SD-MAR-Math samples along with their corresponding grounded reasoning traces. 
This design controls difficulty along both the spatial (multi-country comparison) and temporal (multi-period reasoning) axes, progressively increasing the mathematical complexity and intermediate reasoning steps.

\begin{algorithm}[t]
\footnotesize
\caption{SD-MAR-Math Generation }
\label{alg:MAR_math}
\begin{algorithmic}[1]
\Require World Bank GDP time series $\mathcal{G}$; chart types $\mathcal{C}=\{\text{bar},\text{line},\text{table}\}$; 
levels $\mathcal{L}=\{\text{L1 (Simple)},\text{L2 (Intermediate)},\text{L3 (Complex)}\}$
\State $(\mathcal{K}, T_1, T_2) \gets \Call{SampleCountriesAndIntervals}{\mathcal{G}}$
\State $\mathbf{X}_{T_1}, \mathbf{X}_{T_2} \gets \Call{ExtractGDP}{\mathcal{G}, \mathcal{K}, T_1, T_2}$ \Comment{per-country vectors}
\State $\mu_{k}^{T} \gets \Call{MeanGDP}{\mathbf{X}_{k,T}}$ for all $k\in\mathcal{K}$ and $T \in \{T_1,T_2\}$
\State $L \gets \Call{Level}{\mathcal{L}}$

\If{$L=\mathrm{L1}$}
  \State $q \gets \text{For country k, what is the absolute difference between intervals?}$
  \State $y \gets \left|\mu_{k}^{T_2} - \mu_{k}^{T_1}\right|$
\ElsIf{$L=\mathrm{L2}$}
  \State $q \gets \text{For country k, what is the percentage change between intervals?}$
  \State $y \gets 100 \times \frac{\mu_{k}^{T_2} - \mu_{k}^{T_1}}{\mu_{k}^{T_1}}$
\Else \Comment{L3}
  \State $q \gets \text{Which country has the largest growth rate across intervals?}$
  \State $y \gets \arg\max_{k \in \mathcal{K}} \left( \frac{\mu_{k}^{T_2} - \mu_{k}^{T_1}}{\mu_{k}^{T_1}} \right)$
\EndIf

\State $\mathcal{O} \gets \Call{MakeDistractors}{y, L}$ \Comment{generate distractor options}
\State $c \gets \Call{Chart}{\mathcal{C}}$
\State $v \gets \Call{RenderChart}{\mathbf{X}_{T_1}, \mathbf{X}_{T_2}, \mathcal{K}, c}$ \Comment{matplotlib}
\State $\mathrm{meta} \gets \{\text{countries}= \mathcal{K},\ \text{intervals}= (T_1,T_2),\ \text{level}= L,\ \text{chart type}= c\}$
\State \Return $s \gets \{\text{inputs: } v,\ \text{question: } q,\ \text{choices: } \mathcal{O},\ \text{answer: } y,\ \text{metadata: } \mathrm{meta}\}$
\end{algorithmic}
\end{algorithm}

\paragraph{Factuality Verification and Diversity Analysis}
We conduct automatic filtering and qualitative inspection to examine dataset quality. For SD-MAR-TSE, we apply CLIPScore-based filtering followed by qualitative inspection of sampled examples to examine semantic consistency and potential ambiguity in generated reasoning tasks. For SD-MAR-Math, correctness is guaranteed through deterministic backtracking construction of the reasoning process. We further analyze dataset diversity across change types and reasoning complexity levels. Detailed procedures and statistics are provided in Appendix~\ref{app:dataset_verification}, Appendix~\ref{app:MAR_diversity}, and Appendix~\ref{app:fact_diversity}.

\paragraph{Output Format} 
Both pipelines produce unified samples consisting of multimodal input (images or charts), structured metadata, a question that requires analytical reasoning, four answer choices, and a labeled correct answer. The correct answer serves as a verifiable reward signal for RLVR training. This design ensures that each instance explicitly encodes both \emph{observed differences} and \emph{underlying factors}, making the dataset suitable for training models to perform analytical reasoning across multi-images. More samples of SD-MAR-TSE and SD-MAR-Math are provided in Appendix~\ref{app:dataset_samples}.

\paragraph{From Data to Training: Bridging SD-MAR SDG and RLVR}
SD-MAR provides verifiable supervision through deterministic ground-truth answers, making it directly compatible with reinforcement learning with verifiable rewards (RLVR). However, the multi-step nature of MAR reasoning requires a sequential perceive$\rightarrow$compare$\rightarrow$infer structure, where credit should be allocated non-uniformly across the reasoning trajectory. This introduces additional optimization challenges that are not explicitly considered in standard GRPO. We next describe GRPO-lite, a targeted GRPO configuration designed to exploit the multi-image analytical tasks.

\subsection{SD-MAR RLVR: Reinforcement Learning with Verifiable Rewards}
\label{sec:methods}
To optimize VLMs using reinforcement learning with verifiable rewards (RLVR), we build upon Group Relative Policy Optimization (GRPO) \citep{shao2024deepseekmath}, which updates the policy via a clipped policy gradient objective based on relative comparisons among sampled candidate responses. To address the specific optimization challenges of MAR, we adopt a targeted GRPO configuration referred to as GRPO-lite, which removes components misaligned with the MAR training signal while retaining stable policy updates. The objective follows a clipped policy optimization form, as detailed in Appendix~\ref{sec:grpolite-eq}. GRPO-lite makes three targeted modifications to standard GRPO, each grounded in properties of the MAR setting:

\emph{(i) Removal of KL regularization.}
MAR tasks require models to develop cross-image comparison strategies that may not be explicitly captured by the base policy. Removing KL encourages the policy to explore the novel reasoning patterns that MAR demands while clipped policy updates still prevent catastrophically large shifts.

\emph{(ii) Removal of advantage normalization.} SD-MAR reward distributions are naturally skewed across difficulty levels: Math Level~1 tasks yield high reward density while Level~3 tasks produce sparse rewards. Following Dr.~GRPO~\citep{liu2025understanding}, normalizing advantages across these heterogeneous difficulty levels conflates easy and hard instances, leading to gradient instability. Removing normalization preserves the natural reward structure of SD-MAR's stratified reasoning tasks.



\emph{(iii) Backward Discounted Allocation (BDA).} SD-MAR tasks follow a perceive$\rightarrow$compare$\rightarrow$infer reasoning arc, where early tokens describe individual visual states and later tokens perform cross-image comparison and inference that determine the final answer. This inference pattern parallels sequential decision-making in reinforcement learning, where credit should be assigned across temporally ordered actions~\citep{sutton1998reinforcement}. Motivated by this structure, we adopt BDA, which has been explored in sequence-level reinforcement learning for language generation~\citep{ranzato2015sequence, paulus2017deep, le2025token, tan2025gtpo}. Instead of uniformly distributing a terminal reward across all tokens, BDA applies exponentially higher weight to later reasoning steps while discounting earlier tokens, directly targeting the comparison and inference steps where analytical reasoning occurs.
Given a reasoning sequence of length $L$, the reward assigned to token $t$ is defined as $r_t = R \cdot \frac{\gamma^{L-1-t}}{\sum_{i=0}^{L-1}\gamma^i}$, where $\gamma \in (0,1)$ controls the rate of backward discounting. This formulation allocates greater credit to later reasoning steps that more directly determine the final answer, while still propagating supervision signals to earlier tokens.

\paragraph{GRPO-lite}
\label{sec:grpolite-eq}

We design BDA by assigning token-level rewards \(w_t\) based on the final outcome reward \(R\):
\[
w_t = R \cdot \frac{\gamma^{L-1-t}}{\sum_{i=0}^{L-1}\gamma^i}, \quad \gamma \in (0,1).
\]

The GRPO-lite objective is defined as:
\[
J(\theta)=\mathbb{E}_{q,\{o_i\}}
\left[
\frac{1}{G}\sum_{i=1}^{G}
\frac{1}{|o_i|}
\sum_{t=1}^{|o_i|}
\min\left(r_{i,t}(\theta)A_i,\,
\text{clip}(r_{i,t}(\theta),1-\epsilon_{\text{low}},1+\epsilon_{\text{high}})A_i\right)
\right],
\]

where \(r_{i,t}(\theta)\) is the probability ratio and \(A_i\) is the group-relative advantage computed from sequence-level rewards.

%% file: sec/5_exp_colm_arxiv_v0.tex
\section{Experiments and Results}
\label{sec:exp_results}

\subsection{Research Questions}
\label{sec:research_questions}
To assess the impact of SD-MAR fine-tuning and the effectiveness of GRPO-lite, we investigate the following research questions(RQs):\\
\emph{RQ1}: \textbf{In-Domain Performance} To what extent does fine-tuning on SD-MAR improve multi-image analytical reasoning performance compared to the base models and closed-source baselines? \\
\emph{RQ2}: \textbf{Out-Domain Performance} Does SD-MAR finetuning preserve or improve performance on related vision-language reasoning tasks?\\
\emph{RQ3}: \textbf{Generalization Performance} Does SD-MAR fine-tuned models have catastrophic forgetting on tasks outside the multi-image analytical reasoning domain?

\subsection{Experiment Settings}
\label{sec:exp_setting}

\paragraph{Model Families} 
We conduct experiments on two representative open-source VLM families, InternVL3-8B \citep{zhu2025internvl3} and Qwen2.5-VL-7B \citep{bai2025qwen2}, and additionally report results from the closed-source GPT model (GPT 4.1) as a reference baseline. Detailed training configurations are provided in Appendix~\ref{app:rl_training_config}.

\paragraph{Datasets} 
The SD-MAR data consists of three datasets.  
SD-MAR-TSE (TSE) contains 5,995 examples (5,395 for training, 300 for validation, and 300 for testing).  
SD-MAR-Math (Math) includes 15,000 examples (13,500 for training, 750 for validation, and 750 for testing).  
To study the effect of data diversity during training, we further construct SD-MAR-Mix (Mix), which combines both domains. SD-MAR-Mix contains 20,995 examples (18,895 train, 1,050 validation, 1,050 test).

\paragraph{RL Framework and Reward Design} 
We adopt VeRL \citep{sheng2024hybridflow} as the reinforcement learning framework. Our reward design follows RLVR from DeepSeek-R1~\citep{guo2025deepseek}, which employs a verifiable outcome reward to optimize models based on reasoning accuracy, and output consistency. In our setup, we define two binary evaluation criteria: outcome correctness (\(r_{\text{acc}}\)) and format validity (\(r_{\text{fmt}}\)), with corresponding weights \(w_{\text{acc}} = 0.7\) and \(w_{\text{fmt}} = 0.3\). We place greater emphasis on outcome correctness, while retaining format rewards to mitigate reward sparsity and stabilize training, following common practices in reinforcement learning~\citep{ouyang2022training, guo2025deepseek}.

\paragraph{Evaluation Strategy}
\label{sec:eval_strategy}
To address the research questions in Sec.~\ref{sec:research_questions}, we evaluate fine-tuned models across four settings. \emph{In-domain Evaluation} on SD-MAR directly measures analytical reasoning (Tab.~\ref{tab:MAR_test}, Tab.~\ref{tab:merged_domain_eval}). \emph{Evaluation on Related but Out-of-Domain Tasks} on MMBench~\citep{liu2024mmbench} assesses transfer to related vision–language reasoning tasks (Tab.~\ref{tab:mmbench}) \emph{Evaluation on Generalization Benchmarks} on MME~\citep{zhang2024mme}, MMMU-Pro~\citep{yue2024mmmu}, and MathVista~\citep{lu2023mathvista} tests preservation of general multimodal capability (Tab.~\ref{tab:public_benchmark}). Finally, a \emph{CoT evaluation} using an LLM-as-judge framework~\citep{zheng2023judging} with GPT-4.1 evaluates reasoning quality beyond task accuracy (Tab.~\ref{tab:reasoning_eval}). The LLM-judge evaluation is applied only to the open-source models, while GPT-4.1 is reported as a reference baseline and is not evaluated by the judge. We report evaluation results on related but out-of-domain tasks and generalization benchmarks using the SD-MAR-Mix–trained model in Sec.~\ref{sec:simdomain_eval}, Sec.~\ref{sec:outdomain_eval} and Sec.~\ref{sec:llmjudge}.



\subsection{In-Domain Evaluation: SD-MAR}
\label{sec:indomain_eval}
\paragraph{In-domain Training Analysis: SD-MAR-Mix}
Tab.~\ref{tab:MAR_test} reports in-domain performance on SD-MAR with training on the Mix subset and evaluation on all subsets (Mix, TSE, and Math). GRPO-lite improves average accuracy by 29.0\% for Qwen2.5-VL-7B and 15.4\% for InternVL3-8B, with the largest gains on the Math subset (Qwen-VL\_Mix: 48.8\%$\rightarrow$91.73\%; Intern-VL\_Mix: 40.53\%$\rightarrow$60.40\%). Comparing GRPO-lite with its w/o BDA variant, BDA yields clear gains on Math and Mix by strengthening credit assignment toward later reasoning steps. By allocating higher reward weight to decisive tokens, BDA mitigates reward dilution in sequential reasoning. This effect is weaker for TSE, where reasoning signals are distributed more evenly across description, change detection, and analytical inference.  The larger gains observed for Qwen2.5-VL-7B suggest that GRPO-lite amplifies existing analytical reasoning capacity rather than compensating for weak representations. Qwen2.5-VL-7B’s end-to-end integrated visual–projector–language architecture may facilitate more efficient cross-modal alignment, whereas InternVL3-8B’s hybrid design~\citep{zhu2025internvl3} may limit steerability under trajectory-level reinforcement learning. Additional comparisons with GRPO and DAPO, along with training dynamics, are provided in Appendix~\ref{app:grpolite_grpo_dapo}.


\begin{table}[h!]
    \centering
    \begin{tabular}{l|ccc}
        \toprule
        Model & \scriptsize SD-MAR-Mix\textsubscript{test} & \small SD-MAR-Math\textsubscript{test} & \scriptsize SD-MAR-TSE\textsubscript{test} \\
        \midrule
        \small GPT-4.1 & 80.95\% & 89.33\% & 65.33\% \\
        \midrule
        \small Qwen2.5-VL-7B \textcolor{black}{(base)} & 48.00\% & 48.80\% & 61.33\% \\
        \small Qwen-VL\_Mix GRPO-lite w/o BDA  & 79.52\% & 84.00\% & 80.67\% \\
        \small Qwen-VL\_Mix GRPO-lite  & \textbf{84.95\%} & \textbf{91.73\%} & \textbf{74.33\%} \\
        \midrule
        \small InternVL3-8B \textcolor{black}{(base)} & 47.33\% & 40.53\% & 60.33\% \\
        \small Intern-VL\_Mix GRPO-lite w/o BDA  & 51.14\% & 50.00\% & 60.67\% \\
        \small Intern-VL\_Mix GRPO-lite  & \textbf{52.86\%} & \textbf{60.40\%} & \textbf{62.00\%} \\
        \bottomrule
    \end{tabular}
    \caption{Performance on SD-MAR with training on the Mix subset and evaluation across Mix, Math, and TSE.}
    \label{tab:MAR_test}
\end{table}

\paragraph{In-domain Cross Evaluation: SD-MAR-TSE and SD-MAR-Math}
\label{sec:indomain_cross}
Tab.~\ref{tab:merged_domain_eval} reports same- and cross-domain results on SD-MAR-Math and SD-MAR-TSE. Under same-domain training, GRPO-lite substantially improves Qwen2.5-VL-7B (+45.60\% on Math; +15.34\% on TSE), while gains for InternVL3-8B are smaller and less consistent. Cross-domain transfer is asymmetric: models trained on Math subset exhibit modest generalization to TSE, whereas TSE subset--trained models show limited or negative transfer to Math. This suggests that quantitative analytical reasoning partially transfers to semantic tasks, but the reverse direction remains constrained. Comparing GRPO-lite with its w/o BDA variant, BDA yields clear gains on Math (e.g., +3.07\% for Qwen2.5-VL-7B) but provides limited benefit on TSE. InternVL3-8B shows weaker sensitivity to BDA, potentially reflecting architecture-dependent sensitivity in cross-modal alignment \citep{liu2023visual}.


\begin{table}[h!]
\centering
\begin{tabular}{l|cc|cc}
\toprule
Model & \multicolumn{2}{c}{GRPO-lite (w/o BDA)} & \multicolumn{2}{c}{GRPO-lite} \\
\cmidrule(lr){2-3} \cmidrule(lr){4-5}
& \scriptsize SD-MAR-Math & \scriptsize SD-MAR-TSE & \scriptsize SD-MAR-Math & \scriptsize SD-MAR-TSE \\
\midrule
\small Qwen-VL\_Math & \cellcolor{gray!15}91.33\% & \textit{62.00\%} & \cellcolor{gray!15}94.40\% & \textit{63.33\%} \\
\small Qwen-VL\_TSE  & \textit{36.53\%} & \cellcolor{gray!15}79.67\% & \textit{40.80\%} & \cellcolor{gray!15}76.67\% \\
\small Qwen2.5-VL-7B & 48.80\% & 61.33\% & 48.80\% & 61.33\% \\
\midrule
\small Intern-VL\_Math & \cellcolor{gray!15}42.13\% & \textit{57.67\%} & \cellcolor{gray!15}35.33\% & \textit{57.33\%} \\
\small Intern-VL\_TSE  & \textit{45.73\%} & \cellcolor{gray!15}70.67\% & \textit{33.73\%} & \cellcolor{gray!15}58.67\% \\
\small InternVL3-8B    & 40.53\% & 60.33\% & 40.53\% & 60.33\% \\
\bottomrule
\end{tabular}
\caption{Cross-subset evaluation on SD-MAR. Rows denote models fine-tuned on different subsets (\texttt{\_Math}, \texttt{\_TSE}). Diagonal entries (\cellcolor{gray!15}{shaded}) correspond to same-subset training and evaluation, while off-diagonal entries (\textit{italic}) reflect cross-subset generalization.}
\label{tab:merged_domain_eval}
\end{table}

Fig. \ref{fig:grpo_vs_grpolite_internvl} in Appendix.~\ref{sec:train_dynamics} shows the reward and entropy dynamics of applying BDA within GRPO-lite on Qwen2.5-VL-7B and InternVL3-8B. The results suggest that BDA can lead to reward improvements in some settings, but its effectiveness varies across models; in particular, gains are marginal for InternVL, and instability remains.

\subsection{Evaluation on Related but Out-of-Domain Tasks: MMBench}
\label{sec:simdomain_eval}
Tab.~\ref{tab:mmbench} reports performance on MMBench. For Qwen2.5-VL-7B based models  both GRPO-lite and GRPO-lite (w/o BDA) maintain or slightly improve upon the base model, with an average gain of approximately 1.5\% and the strongest improvement (+1.93\%) achieved under TSE training. For InternVL3-8B, GRPO-lite yields clearer and more consistent gains across settings, with the mixed dataset performing best. Overall, both backbones preserve or modestly enhance similar-domain performance, ranging from about +2\% for Qwen2.5-VL-7B to up to +4\% for InternVL3-8B, indicating that SD-MAR fine-tuning does not compromise related vision–language reasoning ability. Notably, InternVL3-8B fine-tuned with GRPO-lite (w/o BDA) surpasses Ovis2-16B on MMBench despite having fewer parameters, underscoring the effectiveness of the synthetic training signal. A detailed analysis of the differing SD-MAR and MMBench behaviors between Qwen2.5-VL-7B and InternVL3-8B is provided in Appendix~\ref{app:mmbench_eval}. These results address RQ2 by showing that GRPO-lite strengthens in-domain analytical reasoning while sustaining or improving performance on related benchmarks.

\begin{small}
\begin{table}[h!]
\centering
\begin{tabular}{l|c|ccc|ccc}
\toprule
Model & Base & \multicolumn{3}{c}{GRPO-lite (w/o BDA)} & \multicolumn{3}{c}{GRPO-lite} \\
\cmidrule(lr){3-5} \cmidrule(lr){6-8}
& & Mix & Math & TSE & Mix & Math & TSE \\
\midrule
\small Qwen-VL\_Mix   & $\text{82.6\%}^{*}$ & 83.33\% & 83.59\% & 83.67\% & 82.47\% & 82.98\% & \textbf{84.53\%} \\
\small Intern-VL\_Mix  & $\text{81.7\%}^{\dagger}$ & \textbf{86.08\%} & 85.56\% & 85.91\% & \textbf{86.08\%} & 85.65\% & 85.99\% \\
\midrule
\small Ovis2-16B & 85.8\%$^{\ddagger}$ & — & — & — & — & — & — \\
\bottomrule
\end{tabular}
\caption{MMBench evaluation across SD-MAR\_Mix fine-tuned models}
\label{tab:mmbench}
\begin{tablenotes}
    \item {*} results from \citep{bai2025qwen2}; ${\dagger}$ results from \citep{zhu2025internvl3}. $\ddagger$ result for Ovis2-16B on MMBench from the OpenVLM Leaderboard~\citep{duan2024vlmevalkit}.
\end{tablenotes}
\end{table}
\end{small}

\subsection{Evaluation on Generalization Benchmarks: MME, MMMU-Pro, MathVista}
\label{sec:outdomain_eval}

Tab.~\ref{tab:public_benchmark} reports results on three generalized vision-language benchmarks. For both Qwen2.5-VL-7B and InternVL3-8B , GRPO-lite and its w/o BDA variant preserve performance close to their respective baselines, with only minor fluctuations across datasets. Notable gains include Qwen2.5-VL-7B  on MathVista (+3.4\%, 68.2$\rightarrow$72.2) and InternVL3-8B on MME (+0.3\%, 2415.4$\rightarrow$2422.0), while most other changes remain within $\pm$1\% of the baseline.  These findings answer RQ3 by confirming that GRPO-lite fine-tuning on SD-MAR enhances analytical reasoning while maintaining stable out-of-domain generalization.

\begin{footnotesize}
\begin{table}[h!]
\centering
\begin{tabular}{l|l|c|ccc|ccc}
\toprule
Model & BM & Base & \multicolumn{3}{c}{GRPO-lite (w/o BDA)} & \multicolumn{3}{c}{GRPO-lite} \\
\cmidrule(lr){4-6} \cmidrule(lr){7-9}
& & & Mix & Math & TSE & Mix & Math & TSE \\
\midrule
\multirow{3}{*}{Qwen-VL\_Mix}
& $B_1$  & $2347^{*}$ & 2320.86 & 2302.00 & 2340.11 & \textbf{2354.16} & 2328.34 & 2353.61 \\
& $B_2$  & $38.3^{*}$ & 39.03   & 38.90   & 38.46   & 36.47   & 33.89   & \textbf{39.25} \\
& $B_3$  & $68.2^{*}$ & 71.6    & 71.3    & 69.7    & 68.9    & 71.3    & \textbf{72.2} \\
\midrule
\multirow{3}{*}{Intern-VL\_Mix}
& $B_1$  & $2415.4^{\dagger}$ & 2398.06 & 2395.39 & 2398.07 & 2396.93 & 2397.19 & \textbf{2422.03} \\
& $B_2$  & 40.32        & \textbf{40.56} & 40.32 & 40.02 & 40.44 & 40.21 & 40.21 \\
& $B_3$  & $\textbf{71.6}^{\dagger}$ & 69.1 & 68.9 & 69.5 & 69.6 & 68.4 & 70.4 \\
\bottomrule
\end{tabular}
\caption{ Evaluation of SD-MAR-Mix finetuned model on generalized public benchmarks: $BM$ = benchmark,
$B_1$ = MME, $B_2$ = MMMU-Pro, $B_3$ = MathVista.}
\label{tab:public_benchmark}
\begin{tablenotes}
    \item {*} results from \citep{bai2025qwen2}; ${\dagger}$ results from \citep{zhu2025internvl3}.
\end{tablenotes}
\end{table}
\end{footnotesize}

\subsection{LLM-as-Judge Evaluation of Analytical Reasoning Quality under GRPO-lite and GRPO-lite (w/o BDA)}
\label{sec:llmjudge}

To further assess whether SD-MAR and GRPO-lite improve multi-image analytical reasoning beyond task accuracy, we conduct a reasoning quality evaluation using an LLM-as-judge framework~\citep{zheng2023judging}, with GPT-4.1 serving as the evaluator. The judge evaluates generated reasoning traces along five dimensions : \emph{Logic} (coherence and validity of inference steps), \emph{Clarity} (organization and interpretability of the reasoning flow), \emph{Thoroughness} (completeness of multi-step decomposition), \emph{Reflection} (self-consistency and error awareness), and \emph{Readability} (fluency and ease of comprehension). Each dimension is scored on a 0--10 scale, and the overall score is computed as the mean across the five dimensions. To control evaluation variance, all samples are evaluated using a fixed judging prompt with deterministic decoding settings. For each model, the judge evaluates the same set of reasoning outputs generated on the SD-MAR test split. Following common LLM-as-judge practice, the judge receives both the question and the generated reasoning trace but does not have access to the ground-truth label to avoid answer leakage.

We report per-metric improvements relative to the base model, defined as $\Delta = \text{score}_{\text{fine-tuned}} - \text{score}_{\text{base}}$ with detailed statistics reported in Tab. \ref{tab:reasoning_eval}. Representative reasoning examples are provided in Appendix~\ref{app:generation_comparison} Across both model families, GRPO-lite generally improves reasoning quality in logic and clarity while maintaining comparable readability and language consistency, suggesting that SD-MAR training enhances analytical reasoning beyond simple accuracy gains.



\begin{table}[h!]
    \centering    
\begin{tabular}{l|l|ccc|ccc}
\toprule
Model & Metric 
& \multicolumn{3}{c|}{GRPO-lite (w/o BDA)} 
& \multicolumn{3}{c}{GRPO-lite} \\
\cmidrule(lr){3-5} \cmidrule(lr){6-8}
& & Base & Finetuned & $\Delta$ 
& Base & Finetuned & $\Delta$ \\
\midrule
\multirow{6}{*}{Qwen-VL\_Mix}
& Logic        & 7.89 & 8.54 & +0.65 & 8.28 & 8.86 & +0.58 \\
& Clarity      & 8.05 & 8.76 & +0.71 & 8.28 & 9.09 & +0.81 \\
& Thoroughness & 7.91 & 8.52 & +0.61 & 8.35 & 8.71 & +0.36 \\
& Reflection   & 8.02 & 8.40 & +0.38 & 7.59 & 8.34 & +0.75 \\
& Readability  & 9.02 & 9.29 & +0.27 & 8.93 & 9.41 & +0.48 \\
& Overall      & 8.18 & 8.70 & +0.52 & 8.29 & 8.82 & +0.60 \\
\midrule
\multirow{6}{*}{Intern-VL\_Mix}
& Logic        & 7.82 & 8.62 & +0.80 & 8.09 & 8.42 & +0.33 \\
& Clarity      & 7.85 & 8.75 & +0.90 & 8.05 & 8.52 & +0.47 \\
& Thoroughness & 7.73 & 8.71 & +0.98 & 8.23 & 8.19 & -0.04 \\
& Reflection   & 7.83 & 8.65 & +0.82 & 7.41 & 8.24 & +0.83 \\
& Readability  & 8.64 & 9.16 & +0.52 & 8.93 & 8.58 & -0.35 \\
& Overall      & 7.97 & 8.78 & +0.81 & 8.14 & 8.39 & +0.25 \\
\bottomrule
\end{tabular}
    \caption{Reasoning quality evaluation using GPT-4.1 as LLM judge. Scores are reported as Base, Fine-tuned, and $\Delta$, where $\Delta$ denotes the improvement relative to the base model}
    \label{tab:reasoning_eval}
\end{table}

%% file: sec/7_conclusion_colm_sourav_v3.tex
\section{Conclusion}
\label{sec:conclusion}
We propose SD-MAR, a synthetic data framework for enhancing multi-image analytical reasoning in VLMs. SD-MAR introduces a controllable synthetic data generation pipeline that constructs multimodal tasks involving semantic change analysis and quantitative comparison under explicit visual perturbations, enabling targeted and scalable reasoning supervision without manual curation.  We fine-tune Qwen2.5-VL-7B and InternVL3-8B using GRPO-lite with BDA, a targeted GRPO configuration purpose-built for MAR. By removing KL regularization, which constrains the policy to remain close to the base distribution, removing advantage normalization that conflates SD-MAR's stratified difficulty levels, and applying backward-discounted credit assignment to prioritize the decisive comparison and inference steps in multi-image reasoning, GRPO-lite achieves more effective optimization than standard GRPO on MAR tasks. Extensive evaluations across in-domain, out-domain, and more generalized vision-language benchmarks validate both the dataset design and the post-training strategy. These results demonstrate that targeted analytical reasoning enhancement does not compromise broader multimodal capability. Overall, SD-MAR and GRPO-lite provide a scalable approach for improving analytical reasoning through controllable data design and trajectory-level reinforcement learning. Both components are designed for broad reuse: the SD-MAR SDG pipeline is model-agnostic and can generate unbounded training instances for any VLM, while GRPO-lite is task-agnostic and applicable to any RLVR setting with verifiable multi-step reasoning. 

%% file: sec/X_suppl_amazon_arxiv_v1.tex
\setcounter{page}{1}
\renewcommand{\thesubsection}{\Alph{subsection}}  


\subsection{RL Training Configuration}
\label{app:rl_training_config}
We implemented reinforcement learning (RL) using the base models following the Deepseek-R1 configuration \cite{guo2025deepseek}, utilizing the veRL framework \cite{sheng2024hybridflow}. 
For the training configuration, we implemented constant learning rate at $1 \times 10^{-6}$ for the policy model with a rule-based verifiable outcome reward function. In GRPO-lite implementation, we configured the policy with a batch size of 128  and a micro-batch size of 32. During the rollout phase, we collected 128 prompts, each generating 8 sampled responses. The reference KL divergence coefficient was set to 0. We implemented dynamic epsilon parameters with $\epsilon_{\text{low}} = 0.2$ and $\epsilon_{\text{high}} = 0.28$. For ablation setting, vanilla GRPO and DAPO implementation are followed by the algorithms' setting. All experiments were conducted on a cluster of 8 NVIDIA A100 GPUs (80GB each).



\subsection{Training Dynamics Analysis}
\label{sec:train_dynamics}
\begin{figure}[ht]
    \centering
    \begin{minipage}[t]{0.48\linewidth}
        \centering
        \includegraphics[width=\linewidth]{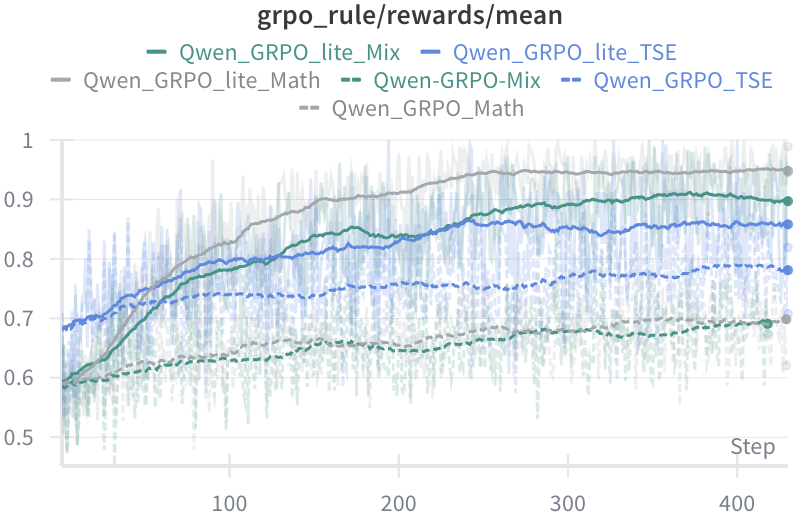}
        \subcaption{Qwen-VL}
        \label{fig:grpo_vs_grpolite_qwen}
    \end{minipage}
    \hfill
    \begin{minipage}[t]{0.48\linewidth}
        \centering
        \includegraphics[width=\linewidth]{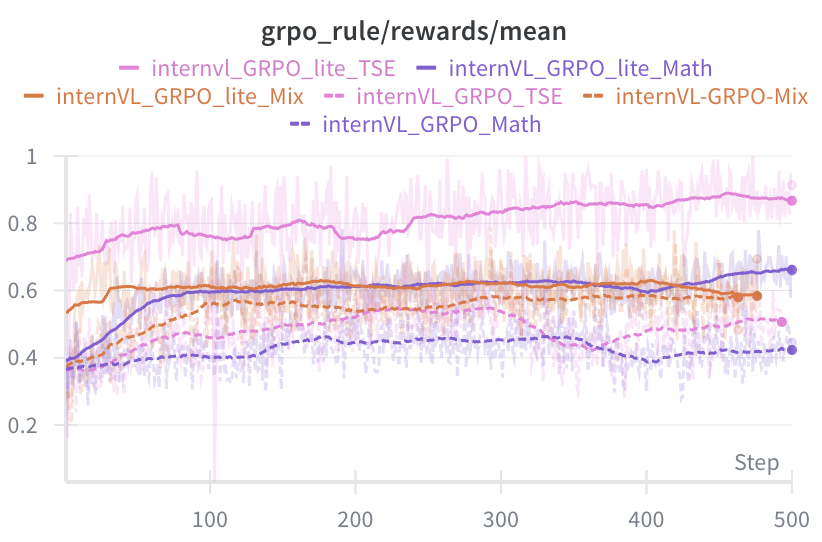}
        \subcaption{Intern-VL}
        \label{fig:grpo_vs_grpolite_internvl}
    \end{minipage}
    \caption{Training dynamics (reward) of GRPO vs.\ GRPO-lite on the mixed dataset for 
    (a) Qwen and (b) InternVL.}
    \label{fig:grpo_vs_grpolite}
\end{figure}

\noindent \textbf{Discussion of GRPO-lite on Two Model Families: Qwen-VL and Intern-VL}
Our experiments reveal distinct patterns in the analytical reasoning capabilities and trainability of Qwen2.5-VL-7B and InternVL3-8B under GRPO-lite fine-tuning on SD-MAR tasks. Qwen2.5-VL-7B consistently outperforms InternVL3-8B after RL fine-tuning across all in-domain test sets. Notably, Qwen2.5-VL-7B with GRPO\_lite achieves 84.95\% on SD-MAR-Mix\textsubscript{test}, surpassing GPT-4.1 by a margin of +4\%, and significantly outperforms the same model's vanilla version (48.00\%), demonstrating the effectiveness of GRPO-lite and its GRPO-lite (w/o BDA) variant.

Training dynamics further corroborate these findings. As shown in Figure~\ref{fig:grpo_vs_grpolite_internvl}, Qwen’s reward curves are smooth and steadily improving across all three sub-tasks, with GRPO-lite variants (especially on SD-MAR-Math) achieving near-saturation around 0.95 reward by step 400. The curves are also more stable with narrower variance bands, indicating consistent policy improvement and high adaptability to analytical reasoning tasks. In contrast, InternVL3-8B exhibits slower convergence and substantially lower final reward ceilings. As visualized in Figure~\ref{fig:grpo_vs_grpolite_internvl}, internVL\_GRPO\_lite\_Math barely exceeds a reward of 0.65, while other domains like TSE show early gains but plateau at lower levels than Qwen. The learning instability, especially in the original GRPO variants (dashed lines), suggests higher variance and brittleness in InternVL’s optimization under sparse or skewed reward distributions.

These discrepancies suggest two key insights. First, the stronger post-training performance of Qwen-VL indicates that its pretraining distribution or architectural inductive biases are better aligned with structured analytical reasoning, enabling more efficient adaptation under trajectory-level reinforcement learning. Second, Intern-VL’s training instability may stem from an overly rigid encoder-decoder alignment, which limits policy adaptation during reward-based learning. Importantly, the success of GRPO-lite variants on both models shows that lightweight RL fine-tuning can enable notable gains without catastrophic forgetting. Our results highlight that analytical reasoning performance is not strictly correlated with model size, but is highly sensitive to pretraining data diversity, fine-tuning protocol, and model adaptability.

\begin{figure}[ht]
    \centering
    \begin{minipage}[t]{0.48\linewidth}
        \centering
        \includegraphics[width=\linewidth]{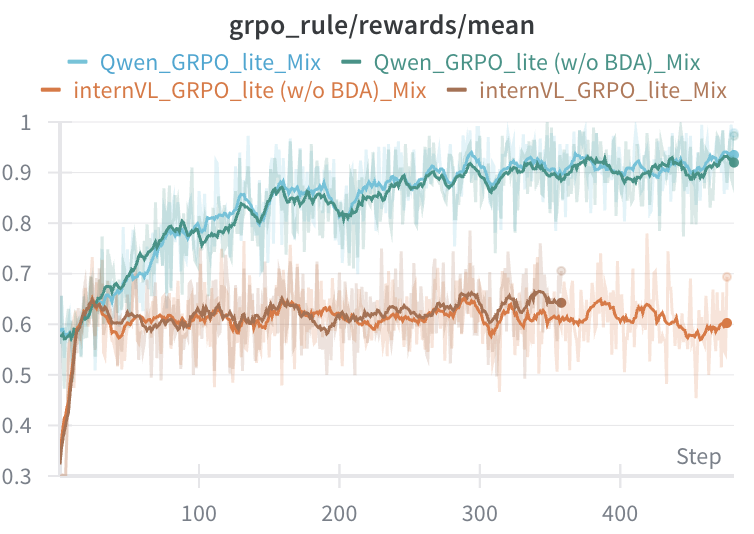}
        \subcaption{Qwen-VL}
        \label{fig:grpo-lite-bda-reward}
    \end{minipage}
    \hfill
    \begin{minipage}[t]{0.48\linewidth}
        \centering
        \includegraphics[width=\linewidth]{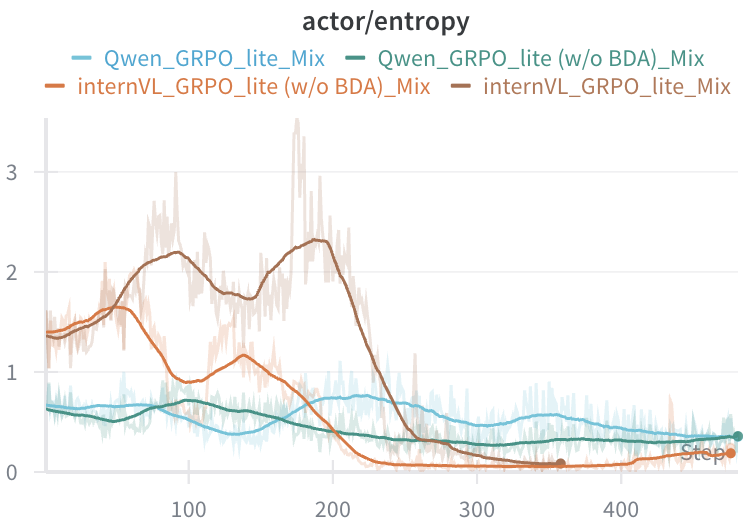}
        \subcaption{Intern-VL}
        \label{fig:grpo-lite-bda-entropy}
    \end{minipage}
    \caption{Training dynamics (reward and entropy) of GRPO-lite (w/o BDA) vs.\ GRPO-lite on the mixed dataset for 
    (a) Qwen-VL and (b) Intern-VL.}
    \label{fig:grpo_vs_grpolite_bda}
\end{figure}

\noindent \textbf{Discussion of GRPO-lite on Two Model Families: Qwen-VL and Intern-VL}
While GRPO-Lite (w/o BDA) already provides an efficient and stable reinforcement fine-tuning approach, integrating BDA further enhances its effectiveness by improving both reward optimization and training stability. As illustrated in Figure~\ref{fig:grpo-lite-bda-reward}, the BDA-enhanced variant achieves consistently higher reward trajectories across training steps, with notably tighter confidence intervals, especially for Qwen-VL, indicating smoother policy updates and more stable learning dynamics. The key contribution of BDA lies in its mechanism that by exponentially weighting later reasoning steps while discounting earlier tokens, it re-balances credit assignment toward longer-horizon, globally coherent behaviors, which helps mitigate the variance induced by short-sighted or noisy feedback common in sparse analytical reasoning rewards.
For Qwen-VL, this results in a gradual but steady reduction in entropy (Figure~\ref{fig:grpo-lite-bda-entropy}), signifying confident policy refinement. In contrast, Intern-VL’s entropy curve under BDA reveals an initially high-exploration regime followed by premature collapse, suggesting that while BDA improves reward alignment, its benefits are modulated by the model's intrinsic policy expressiveness and pretraining alignment. These dynamics underscore that BDA’s temporal smoothing of reward signals not only enhances convergence but also increases robustness to skewed or heavy-tailed reward distributions—conditions frequently encountered in real-world reasoning datasets like SD-MAR.

\subsection{Ablation}
\label{app:ablation}
\subsubsection{GRPO-lite vs GRPO and DAPO}
\label{app:grpolite_grpo_dapo}

We analyze the effect of different training strategies on SD-MAR finetuning using W\&B trajectories, comparing GRPO-lite against GRPO and DAPO. This setup allows us to evaluate training efficiency, optimization behavior, exploration dynamics, and resource utilization without requiring additional decoding.

\textbf{Training efficiency.}
GRPO-lite reaches high rewards early in training with lower per-step cost, reflecting its lightweight design and relaxed regularization. GRPO converges more slowly due to stronger regularization and conservative policy updates. DAPO exhibits moderate convergence speed, benefiting from advantage filtering while maintaining stable updates.

\textbf{Optimization dynamics.}
GRPO-lite shows higher variance in training curves, particularly during early exploration, reflecting its removal of KL regularization and advantage normalization. GRPO produces more conservative updates with smoother trajectories. DAPO yields relatively smooth reward and advantage curves due to its filtering mechanism.

\textbf{Performance.}
GRPO-lite achieves strong reward growth and competitive final performance across the Mix, Math, and TSE datasets. GRPO attains similar performance but requires more training steps to reach comparable reward levels. DAPO also achieves competitive rewards with smoother optimization dynamics.

\textbf{Exploration.}
Entropy trajectories indicate that GRPO-lite maintains stronger exploration early in training, while GRPO gradually reduces entropy through more conservative updates. DAPO maintains relatively stable entropy and response lengths due to advantage filtering.

\textbf{Resource use.}
GRPO-lite is the most lightweight configuration, requiring fewer updates and reduced computational overhead. GRPO incurs higher compute cost due to additional normalization and regularization. DAPO lies between the two in computational complexity.

\textbf{Summary.}
Overall, GRPO-lite emphasizes efficient exploration and fast reward improvement with minimal training overhead, while GRPO and DAPO provide smoother optimization dynamics through stronger regularization and filtering. These patterns are consistent across both Qwen-VL and Intern-VL models, with TSE remaining the most challenging domain.

\begin{figure*}[ht]
  \centering
  \begin{minipage}[t]{0.24\linewidth}
    \centering
    \includegraphics[width=\linewidth]{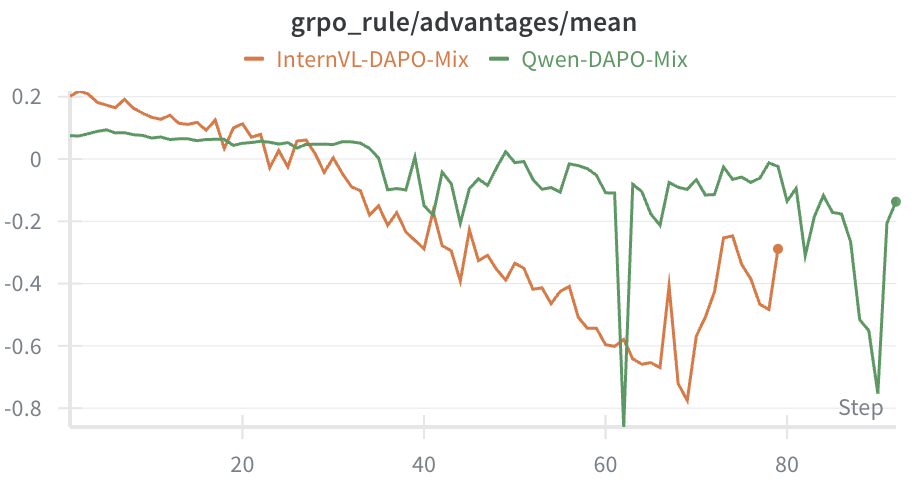}
    \subcaption{DAPO - Advantages}
    \label{fig:dapo_adv}
  \end{minipage}\hfill
  \begin{minipage}[t]{0.24\linewidth}
    \centering
    \includegraphics[width=\linewidth]{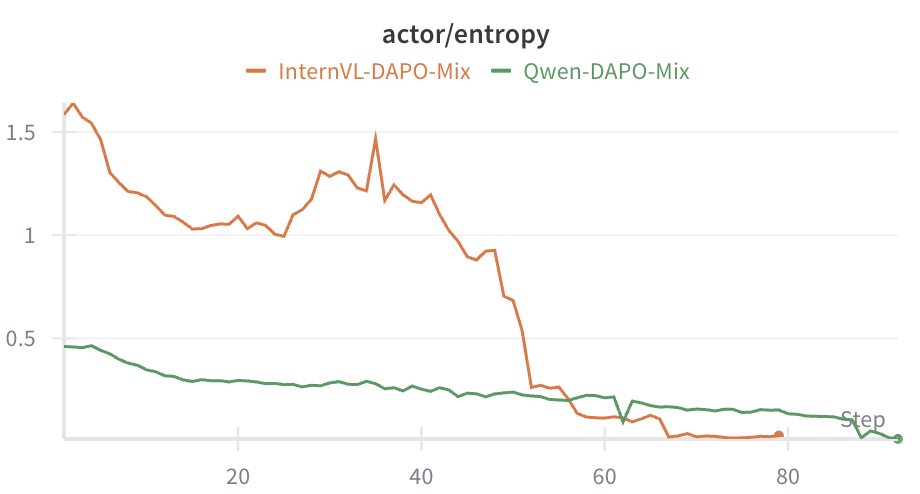}
    \subcaption{DAPO - Entropy}
    \label{fig:dapo_entropy}
  \end{minipage}\hfill
  \begin{minipage}[t]{0.24\linewidth}
    \centering
    \includegraphics[width=\linewidth]{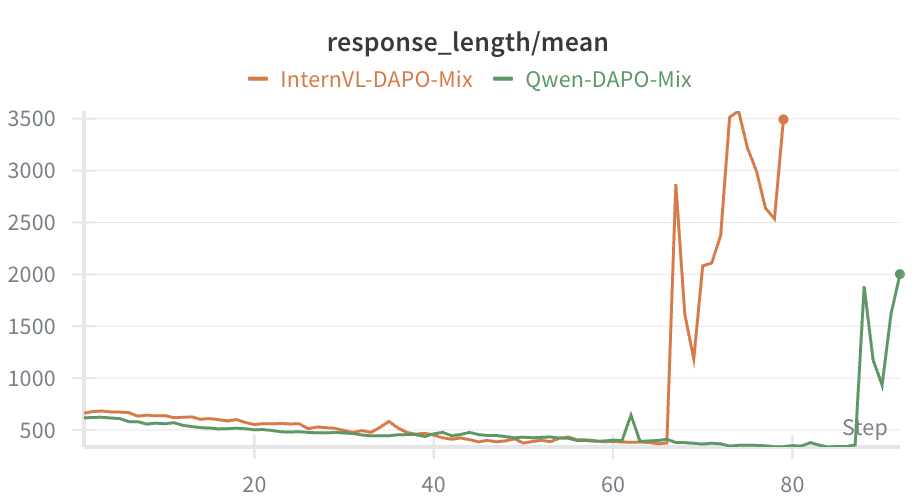}
    \subcaption{DAPO - CoT length}
    \label{fig:dapo_cot_len}
  \end{minipage}\hfill
  \begin{minipage}[t]{0.24\linewidth}
    \centering
    \includegraphics[width=\linewidth]{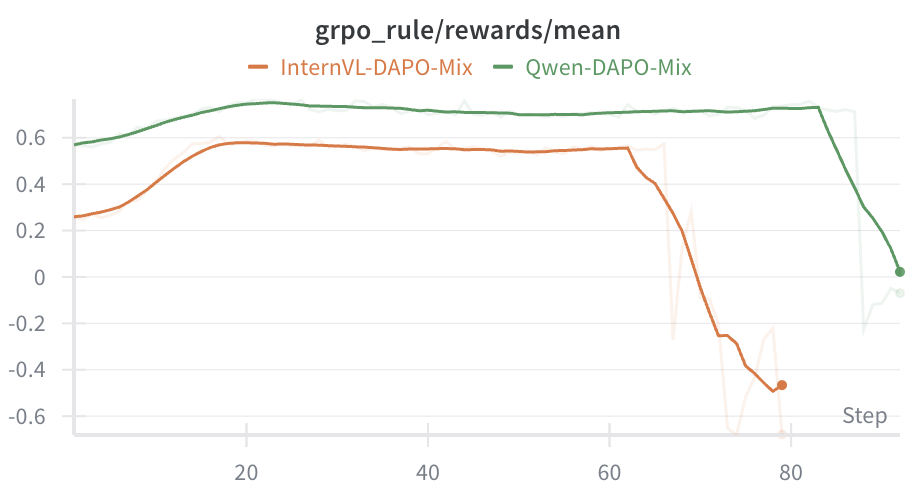}
    \subcaption{DAPO - Rewards}
    \label{fig:dapo_rewards}
  \end{minipage}

  \vspace{0.8em} 

  \begin{minipage}[t]{0.24\linewidth}
    \centering
    \includegraphics[width=\linewidth]{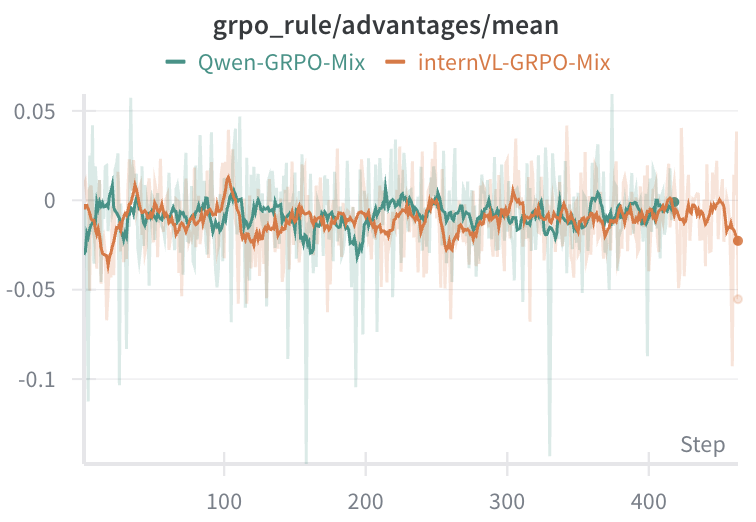}
    \subcaption{GRPO - Advantages}
    \label{fig:grpo_adv}
  \end{minipage}\hfill
  \begin{minipage}[t]{0.24\linewidth}
    \centering
    \includegraphics[width=\linewidth]{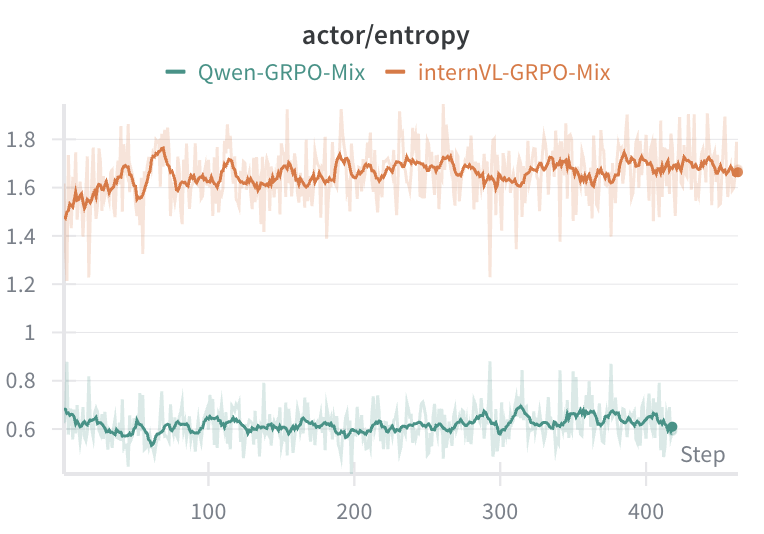}
    \subcaption{GRPO - Entropy}
    \label{fig:grpo_entropy}
  \end{minipage}\hfill
  \begin{minipage}[t]{0.24\linewidth}
    \centering
    \includegraphics[width=\linewidth]{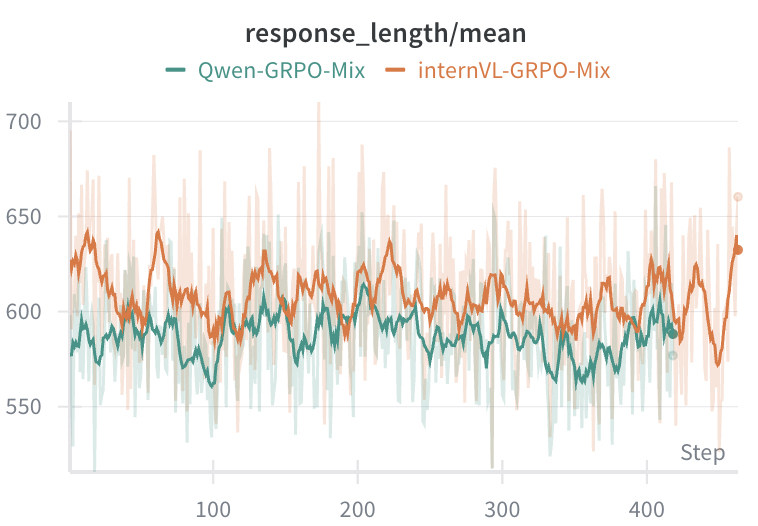}
    \subcaption{GRPO - CoT length}
    \label{fig:grpo_cot_len}
  \end{minipage}\hfill
  \begin{minipage}[t]{0.24\linewidth}
    \centering
    \includegraphics[width=\linewidth]{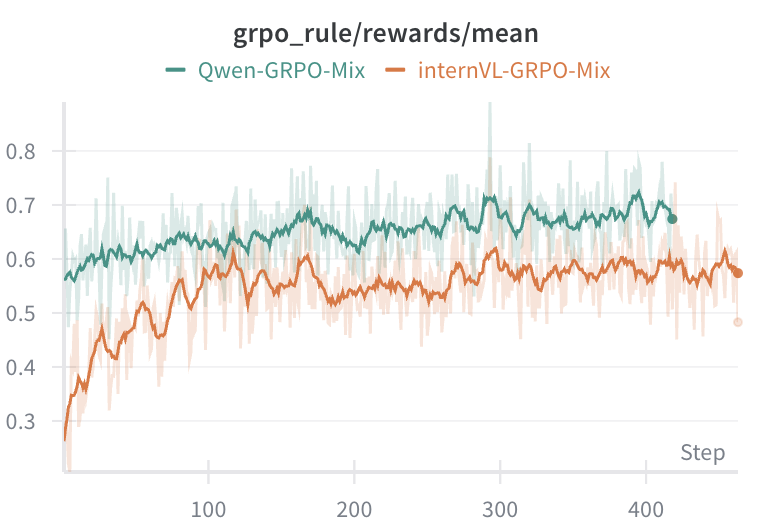}
    \subcaption{GRPO - Rewards}
    \label{fig:grpo_rewards}
  \end{minipage}

  \vspace{0.8em} 

  \begin{minipage}[t]{0.24\linewidth}
    \centering
    \includegraphics[width=\linewidth]{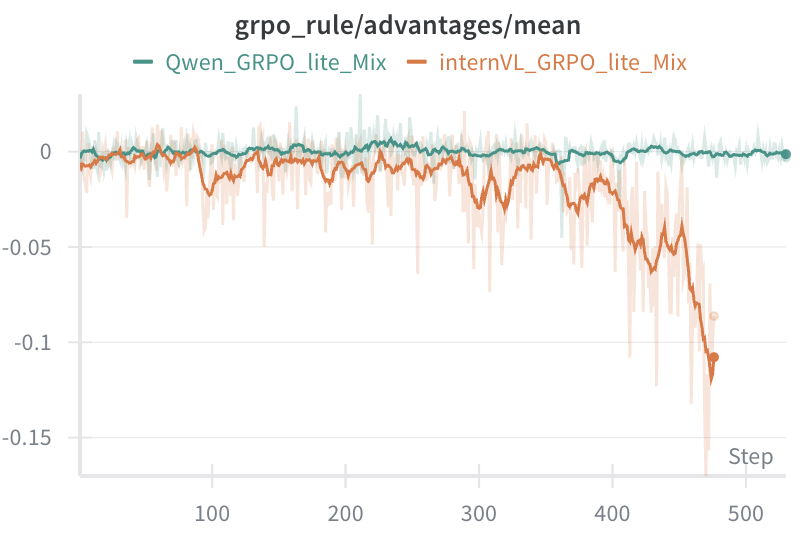}
    \subcaption{GRPO-lite - Advantages}
    \label{fig:grpo_lite_adv}
  \end{minipage}\hfill
  \begin{minipage}[t]{0.24\linewidth}
    \centering
    \includegraphics[width=\linewidth]{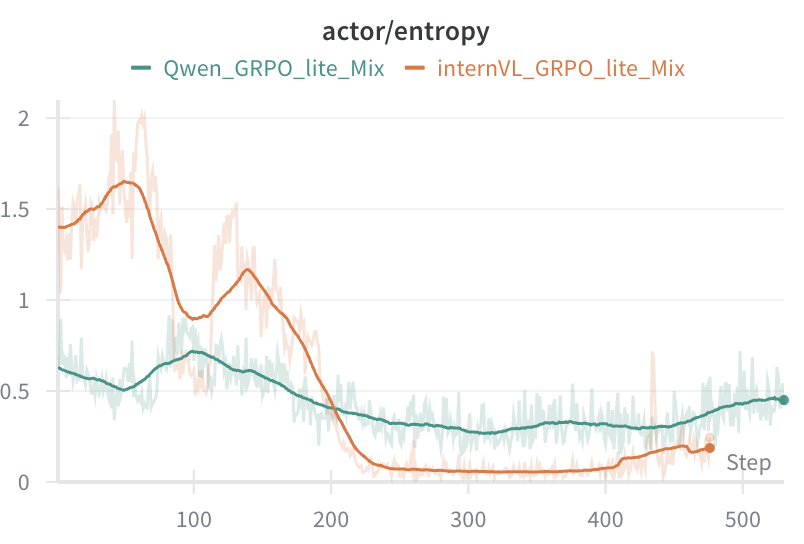}
    \subcaption{GRPO-lite - Entropy}
    \label{fig:grpo_lite_entropy}
  \end{minipage}\hfill
  \begin{minipage}[t]{0.24\linewidth}
    \centering
    \includegraphics[width=\linewidth]{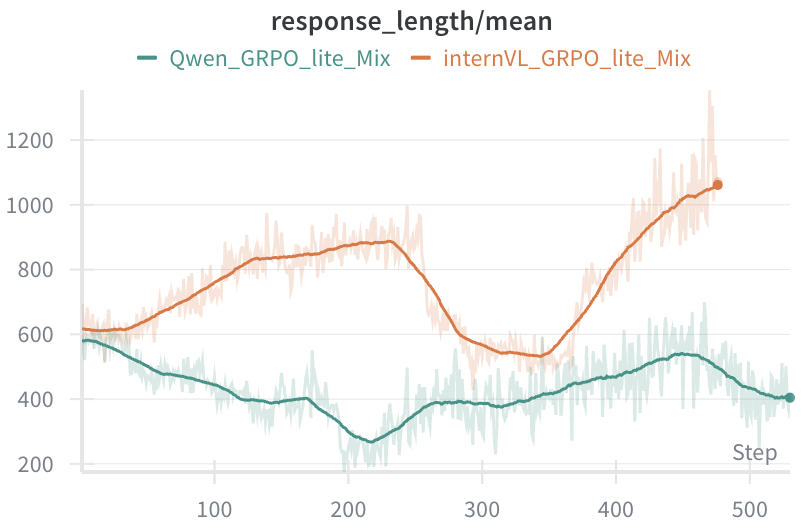}
    \subcaption{GRPO-lite - CoT length}
    \label{fig:grpo_lite_cot_len}
  \end{minipage}\hfill
  \begin{minipage}[t]{0.24\linewidth}
    \centering
    \includegraphics[width=\linewidth]{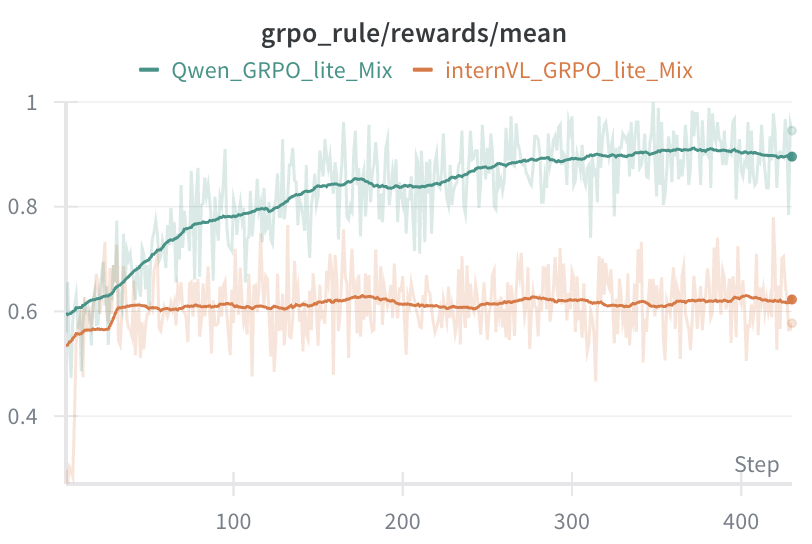}
    \subcaption{GRPO-lite - Rewards}
    \label{fig:grpo_lite_rewards}
  \end{minipage}
  \caption{Ablation dynamics on the mixed dataset for DAPO, GRPO, and GRPO-lite. \textcolor{orange}{Orange curves correspond to InternVL}, and \textcolor{teal}{green curves correspond to Qwen-VL}. Each row shows the training curves of (a) Advantages, (b) Entropy, (c) Response length, and (d) Rewards.}
  \label{fig:ablation_all}
\end{figure*}

\subsubsection{$\gamma$ Comparison under BDA Setting}
\label{app:bda_gamma}

\begin{figure}
    \centering
    \includegraphics[width=0.95\linewidth]{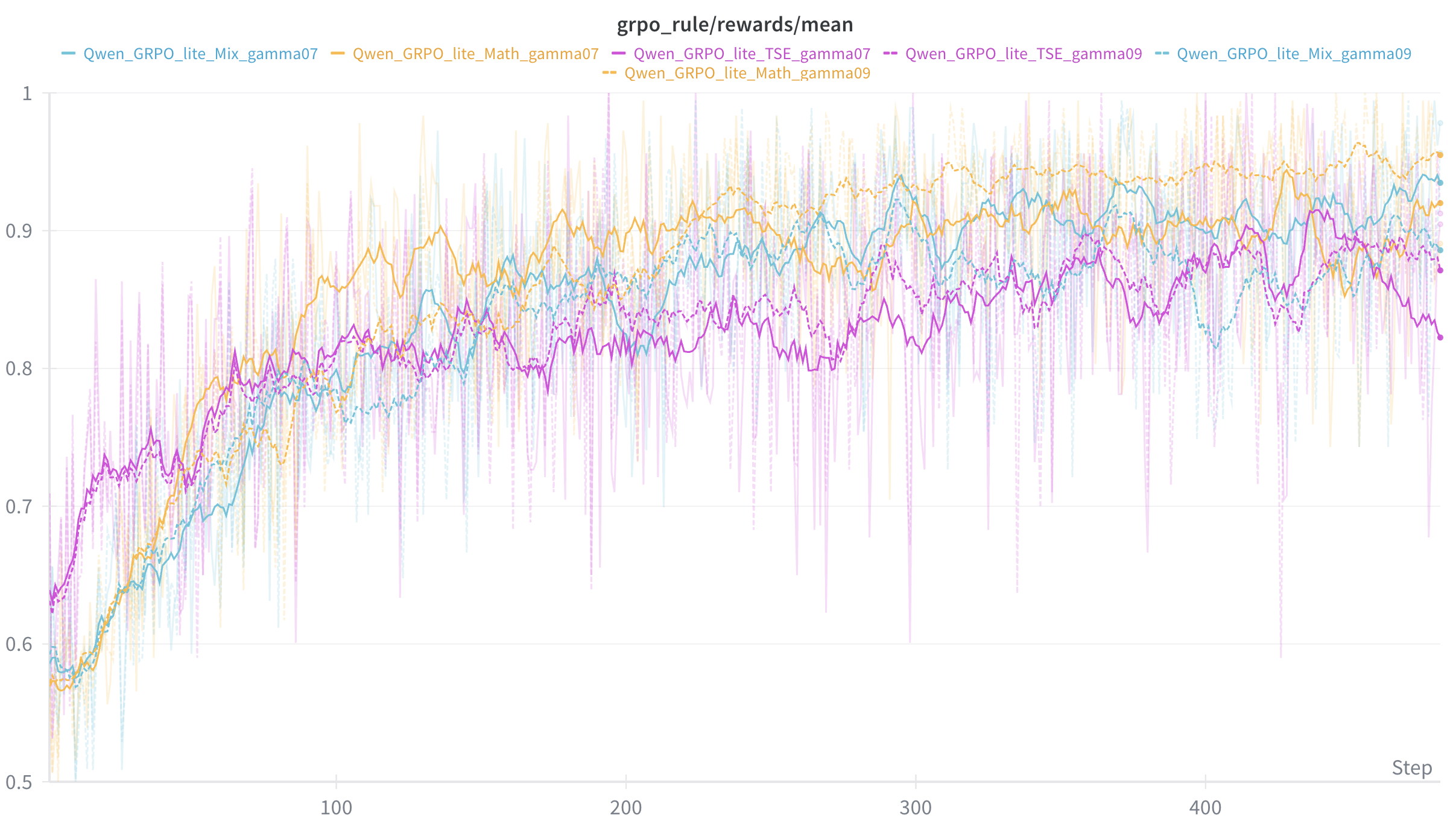}
    \caption{Unified view of $\gamma$ ablation across SD-MAR\_Math, SD-MAR\_Mix, and SD-MAR\_TSE, highlighting how the coefficient influences model learning dynamics. Light blue denotes SD-MAR\_Mix, yellow denotes SD-MAR\_Math, and purple denotes SD-MAR\_TSE. Solid lines indicate $\gamma = 0.7$, while dashed lines indicate $\gamma = 0.9$.}
    \label{fig:main_gamma}
\end{figure}

\begin{figure*}[ht]
  \centering
  \begin{minipage}[t]{0.32\linewidth}
    \centering
    \includegraphics[width=\linewidth]{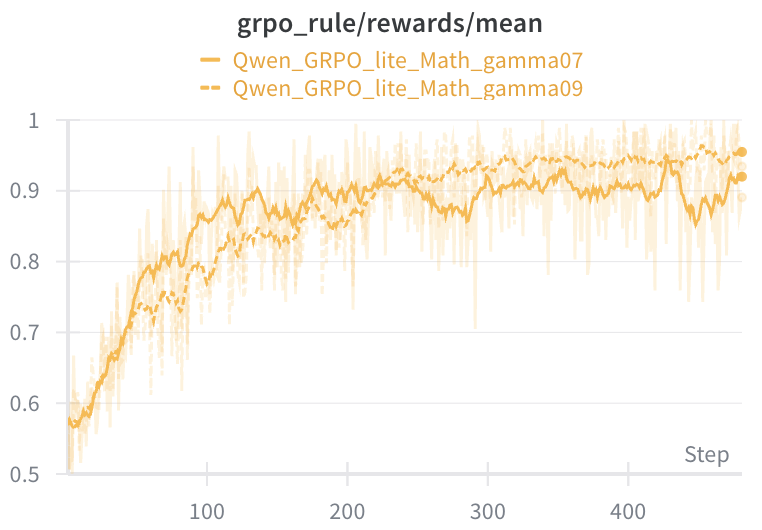}
    \subcaption{SD-MAR\_Math}
    \label{fig:math_gamma}
  \end{minipage}\hfill
  \begin{minipage}[t]{0.32\linewidth}
    \centering
    \includegraphics[width=\linewidth]{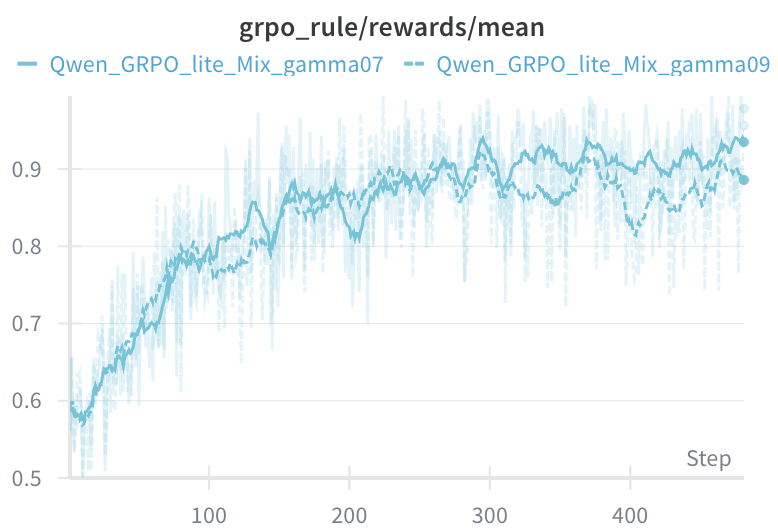}
    \subcaption{SD-MAR\_Mix}
    \label{fig:mix_gamma}
  \end{minipage}\hfill
  \begin{minipage}[t]{0.32\linewidth}
    \centering
    \includegraphics[width=\linewidth]{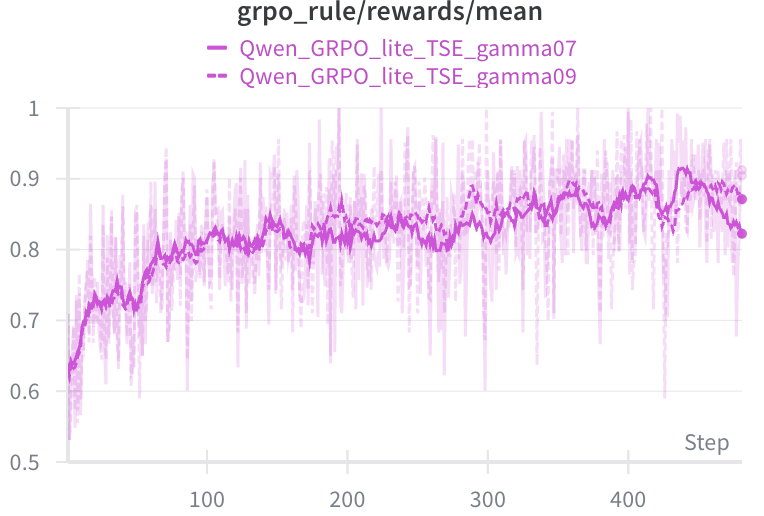}
    \subcaption{SD-MAR\_TSE}
    \label{fig:tse_gamma}
  \end{minipage}
  \caption{Ablation on the $\gamma$ coefficient across different SD-MAR tracks. Each panel shows the training dynamics for (a) SD-MAR\_Math, (b) SD-MAR\_Mix, and (c) SD-MAR\_TSE, highlighting how varying $\gamma$ influences convergence and stability. Light blue denotes SD-MAR\_Mix, yellow denotes SD-MAR\_Math, and purple denotes SD-MAR\_TSE. Solid lines indicate $\gamma = 0.7$, while dashed lines indicate $\gamma = 0.9$.}
  \label{fig:gamma_ablation}
\end{figure*}

\paragraph{Ablation on $\gamma$.}
Fig.~\ref{fig:main_gamma} examines the impact of varying the $\gamma$ coefficient across the three SD-MAR tracks. We compare $\gamma=0.7$ and $\gamma=0.9$ settings under GRPO-lite. On SD-MAR\_Math Fig. ~\ref{fig:math_gamma}, both settings achieve strong convergence, with $\gamma=0.9$ yielding higher and more stable reward trajectories. For SD-MAR\_Mix Fig.~\ref{fig:mix_gamma}, the two settings perform comparably, with $\gamma=0.7$ showing a modest advantage in the later stages of training. By contrast, on SD-MAR\_TSE Fig.~\ref{fig:tse_gamma} the curves are less consistent across steps, and the gains from a larger $\gamma$ are marginal, suggesting that temporal–semantic reasoning is less sensitive to this parameter.  these results indicate that a moderately larger $\gamma$ (e.g., 0.9) enhances performance on math-intensive tasks, while its effect is less pronounced for sematic reasoning. In more diversified settings such as SD-MAR\_Mix, $\gamma=0.7$ results better outcomes. This aligns with our assumption that SD-MAR\_Math tasks follow a structured logical chain, visual understanding, comparison, and calculation, making them more sensitive to the $\gamma$ setting.

\subsection{Benchmark Evaluations}
We report detailed results on MMBench and MME subtasks, with base model references taken from \citet{duan2024vlmevalkit}. For MMBench, percentage symbols are omitted for readability. For MME, we present both perception and cognition scores for base and fine-tuned models.

\subsubsection{MMBench}
\label{app:mmbench_eval}
The contrasting behavior of Intern-VL across SD-MAR and MMBench can be explained by differences in both task demands and model architecture.
On SD-MAR, performance is driven primarily by analytical reasoning across sequential steps, especially in Math, where correctness depends on accurate numerical aggregation at the final reasoning stage. Here, Qwen-VL has a clear advantage due to its end-to-end visual–projector-language design, which tightly couples visual features with language reasoning. GRPO-lite fine-tuning boosts this alignment, resulting large in-domain gains. By contrast, Intern-VL adopts a hybrid structure (Intern-developed vision tower with a Qwen-based language model~\citep{zhu2025internvl3}), which reduces cross-modal alignment efficiency and limits the benefits of reasoning-focused fine-tuning.
In MMBench, the evaluation emphasizes general multimodal perception and recognition tasks, where reasoning depth is less critical. Under these conditions, Intern-VL’s strong vision backbone becomes an advantage, and the hybrid design does not penalize performance as much. As a result, fine-tuned Intern-VL achieves consistent improvements and, in some cases, surpasses larger models such as Ovis2-16B, despite its weaker performance on SD-MAR.

\noindent \textbf{Qwen-VL}
\begin{table}[h!]
    \centering
    \resizebox{\textwidth}{!}{
    \begin{tabular}{lccc|cc|ccc}
        \toprule
        \textbf{Model} & \multicolumn{3}{c}{\textbf{Attribute Reasoning}} & \multicolumn{2}{c}{\textbf{Logic Reasoning}} & \multicolumn{3}{c}{\textbf{Relation Reasoning}} \\
        \cmidrule(lr){2-4} \cmidrule(lr){5-6} \cmidrule(lr){7-9}
        & Physical Property & Function & Identity & Structuralized & Future & Social & Physical & Nature \\ 
        & Reasoning & Reasoning & Reasoning & Image-Text & Prediction & Relation & Relation & Relation \\
        \midrule
        $\text{Qwen2.5-VL-7B}^{*}$ & 62.0 & 95.6 & 96.1 & 67.9 & 77.3 & 91.2 & 56.0 & 89.1 \\
        \midrule
        $\text{SD-MAR\_Math}$ & 68.0 & 92.4 & 97.7 & 76.9 & 62.5 & 88.3 & 58.3 & 77.1 \\
        $\text{SD-MAR\_TSE}$ &72.0  &92.4 &97.7 & 80.8 & 70.0 & 90.7 & 58.3 & 81.3 \\
        $\text{SD-MAR\_Mix}$ & 62.7 & 92.40 & 95.6 & 76.9 & 65.0 & 90.7 & 58.3 & 77.1 \\
        \bottomrule
    \end{tabular}
    }
    \caption{Detailed Breakdown of Qwen-VL on MMBench Reasoning Subtasks}
    \label{tab:reasoning_detailed_qwen}
    \vspace{1ex}
    \begin{minipage}{\textwidth}
        \footnotesize \textit{Note.} $^{*}$ Results from \cite{duan2024vlmevalkit}.
    \end{minipage}
\end{table}

\begin{table}[h!]
    \centering
    \resizebox{\textwidth}{!}{
    \begin{tabular}{lccccc|cccc|ccc}
        \toprule
        \textbf{Model} & \multicolumn{5}{c}{\textbf{Coarse Perception}} & \multicolumn{4}{c}{\textbf{Fine-grained Perception}} & \multicolumn{3}{c}{\textbf{Relational/Other Perception}} \\
        \cmidrule(lr){2-6} \cmidrule(lr){7-10} \cmidrule(lr){11-13}
        & Image & Image & Image & Image & Image & Object & Attribute & Celebrity & OCR & Spatial & Attribute & Action \\ 
        & Style & Scene & Emotion & Quality & Topic & Localization & Recognition & Recognition & & Relationship & Comparison & Recognition \\
        \midrule
        $\text{Qwen2.5-VL-7B}^{*}$ & 81.5 & 90.3 & 78.9 & 60.5 & 97.8 & 81.9 & 91.0 & 91.2 & 95.6 & 66.7 & 84.0 & 86.3 \\
        \midrule
        $\text{SD-MAR\_Math}$ & 94.3 & 99.0 & 82.0 & 64.1 & 91.6 & 65.4 & 92.5 & 94.9 & 92.3 & 46.6 & 86.3 & 92.6 \\
        $\text{SD-MAR\_TSE}$ & 94.3 & 98.1 & 82.0 & 67.9 & 94.4 & 64.2 & 94.6 &97.0  & 94.9 & 51.1 & 88.6 & 92.6 \\
        $\text{SD-MAR\_Mix}$ & 90.6 & 98.1 & 78.0 & 64.1 & 88.9 & 64.2 & 97.3 & 96.0 & 92.3 & 51.1 & 86.4 & 92.6 \\
        \bottomrule
    \end{tabular}
    }
    \caption{Detailed Breakdown of Qwen-VL on MMBench Perception Subtasks}
    \label{tab:perception_detailed_qwen}
    \begin{minipage}{\textwidth}
        \footnotesize \textit{Note.} $^{*}$ Results from \cite{duan2024vlmevalkit}.
    \end{minipage}
\end{table}

\noindent \textbf{Intern-VL}
\begin{table}[h!]
    \centering
    \resizebox{\textwidth}{!}{
    \begin{tabular}{lccc|cc|ccc}
        \toprule
        \textbf{Model} 
        & \multicolumn{3}{c}{\textbf{Attribute Reasoning}} 
        & \multicolumn{2}{c}{\textbf{Logic Reasoning}} 
        & \multicolumn{3}{c}{\textbf{Relation Reasoning}} \\
        \cmidrule(lr){2-4} \cmidrule(lr){5-6} \cmidrule(lr){7-9}
        & Physical Property & Function & Identity & Structuralized & Future & Social & Physical & Nature \\ 
        & Reasoning & Reasoning & Reasoning & Image-Text & Prediction & Relation & Relation & Relation \\
        \midrule
        $\text{InternVL3-8B}^{*}$ & 64.6 & 88.9 & 98.7 & 67.0 & 77.3 & 86.8 & 72.0 & 88.0 \\ 
        \midrule
        $\text{SD-MAR\_Math}$ & 72.0 & 89.9 & 100.0 & 80.8 & 65.0 & 93.0 & 83.3 & 91.7 \\
        $\text{SD-MAR\_TSE}$  & 72.0 & 89.9 & 100.0 & 84.6 & 65.0 & 93.0 & 83.3 & 91.7 \\
        $\text{SD-MAR\_Mix}$  & 73.3 & 89.9 & 100.0 & 83.3 & 62.5 & 95.3 & 83.3 & 91.7 \\
        \bottomrule
    \end{tabular}
    }
    \vspace{1ex}
    \caption{Detailed Breakdown of Intern-VL on MMBench Reasoning Subtasks}
    \label{tab:reasoning_detailed_internvl}
    \begin{minipage}{\textwidth}
        \footnotesize \textit{Note.} $^{*}$ Results from \cite{duan2024vlmevalkit}.
    \end{minipage}
\end{table}

\begin{table}[h!]
    \centering
    \resizebox{\textwidth}{!}{
    \begin{tabular}{lccccc|cccc|ccc}
        \toprule
        \textbf{Model} 
        & \multicolumn{5}{c}{\textbf{Coarse Perception}} 
        & \multicolumn{4}{c}{\textbf{Fine-grained Perception}} 
        & \multicolumn{3}{c}{\textbf{Relational / Other Perception}} \\
        \cmidrule(lr){2-6} \cmidrule(lr){7-10} \cmidrule(lr){11-13}
        & Image & Image & Image & Image & Image & Object & Attribute & Celebrity & OCR & Spatial & Attribute & Action \\ 
        & Style & Scene & Emotion & Quality & Topic & Localization & Recognition & Recognition & & Relationship & Comparison & Recognition \\
        \midrule
        $\text{InternVL3-8B}^{*}$ & 80.4 & 88.9 & 82.2 & 58.9 & 97.8 & 81.9 & 94.4 & 83.7 & 96.7 & 61.3 & 82.7 & 93.2 \\
        \midrule
        $\text{SD-MAR\_Math}$ & 98.1 & 99.0 & 82.0 & 67.9 & 88.9 & 75.3 & 97.3 & 90.9 & 94.9 & 57.8 & 77.3 & 92.6 \\
        $\text{SD-MAR\_TSE}$  & 98.1 & 99.0 & 82.0 & 67.9 & 88.9 & 75.3 & 97.3 & 91.9 & 94.9 & 57.8 & 77.3 & 92.6 \\
        $\text{SD-MAR\_Mix}$  & 98.1 & 99.0 & 82.0 & 67.9 & 88.9 & 75.3 & 97.3 & 91.9 & 94.9 & 55.6 & 81.8 & 92.6 \\
        \bottomrule
    \end{tabular}
    }
    \caption{Detailed Breakdown of Intern-VL on MMBench Perception Subtasks}
    \label{tab:perception_detailed_internvl}
    \begin{minipage}{\textwidth}
        \footnotesize \textit{Note.} $^{*}$ Results from \cite{duan2024vlmevalkit}.
    \end{minipage}
\end{table}

\subsubsection{MME}
On MME, Qwen-VL shows stable perception performance across finetuning variants, with all scores remaining close to the base model (1698.1). Notably, cognition improves substantially under all SD-MAR settings, with the mixed dataset achieving the strongest gain (664.3 vs. 613.9), suggesting that SD-MAR fine-tuning enhances reasoning-oriented abilities without sacrificing perceptual skills.
For Intern-VL, perception scores remain consistently high, with only minor fluctuations around the base model (1748.4). More importantly, cognition benefits from SD-MAR fine-tuning, with the TSE variant achieving the highest score (681.4 vs. 673.6 base). These results indicate that SD-MAR finetuning transfers effectively across architectures, strengthening cognitive reasoning while preserving perceptual competence.
Overall, the MME results directly answer RQ3, demonstrating that SD-MAR fine-tuning improves analytical reasoning capabilities without incurring catastrophic forgetting on out-of-domain tasks, thereby maintaining strong generalization.

\noindent \textbf{Qwen-VL}
\footnotesize{
\begin{table}[h!]
    \centering
    \begin{tabular}{lc|c}
        \toprule
        \multirow{2}{*}{\textbf{Model}} & \multicolumn{2}{c}{\textbf{Out-of-domain (MME\_test$^{\dagger}$) $\uparrow$}} \\
        \cmidrule(lr){2-3}
        & Perception & Cognition \\
        \midrule
        $\text{Qwen2.5-VL-7B}^{*}$ & 1698.1 & 613.9 \\
        $\text{SD-MAR\_Math}$ & 1677.27 & 651.07 \\
        $\text{SD-MAR\_TSE}$  & 1707.90 & 645.71 \\
        $\text{SD-MAR\_Mix}$  & 1689.88 & 664.29 \\
        \bottomrule
    \end{tabular}
    \caption{Out-of-Domain Evaluation for Qwen-VL}
    \label{tab:ablation_qwen}
    \begin{minipage}{\textwidth}
        \footnotesize \textit{Note.} $^{*}$ Results from \cite{duan2024vlmevalkit}.  
        $^{\dagger}$ MME scores are reported as Perception and Cognition.
        \end{minipage}
\end{table}
}


\noindent \textbf{Intern-VL}
\footnotesize{
\begin{table}[h!]
    \centering
    \begin{tabular}{lc|c}
        \toprule
        \multirow{2}{*}{\textbf{Model}} & \multicolumn{2}{c}{\textbf{Out-of-domain (MME\_test$^{\dagger}$) $\uparrow$}} \\
        \cmidrule(lr){2-3}
        & Perception & Cognition   \\
        \midrule
        $\text{InternVL3-8B}^{*}$ & 1748.4 & 673.6  \\
        $\text{SD-MAR\_Math}$ & 1721.8 & 675.4 \\
        $\text{SD-MAR\_TSE}$ & 1740.6 & 681.4 \\
        $\text{SD-MAR\_Mix}$ & 1730.5 & 666.4 \\
        \bottomrule
    \end{tabular}
    \caption{Out-of-Domain Evaluation for Intern-VL}
    \label{tab:ablation_internvl}
    \begin{minipage}{\textwidth}
        \footnotesize \textit{Note.} $^{*}$ Results from \cite{duan2024vlmevalkit}.  
        $^{\dagger}$ MME scores are reported as Perception and Cognition.  
    \end{minipage}
\end{table}
}

\subsection{Dataset}
\label{app:datasets}

\subsubsection{Caption Generator Seeds}
\label{app:generator_seeds}

To support the construction of SD-MAR tasks, we design structured generator seeds that define a space of controllable visual transformations and attribute-level perturbations. These seeds encode domain knowledge across three structured dimensions: \textbf{temporal}, \textbf{spatial}, and \textbf{emotional}. Each generator specifies (i) a taxonomy of structured changes, (ii) a set of aligned underlying factors, and (iii) sampling functions that instantiate concrete scenarios. This design enables reproducible, controllable, and extensible dataset generation for systematic analytical reasoning.

\paragraph{Temporal Transformations.}  
\label{app:temporal_seed}
Temporal seeds capture transformations over time, such as growth, decay, construction, or social change. Each entry is aligned with its underlying mechanism (e.g., ``plant life cycle'' $\rightarrow$ ``genetic programming''). These specifications are used to instantiate paired captions describing an initial and final stage of change.  

\noindent \textbf{Temporal Transformation Generator Seeds}
\begin{lstlisting}[
    language=json,
    frame=single,
    basicstyle=\ttfamily\footnotesize,
    breaklines=true,
    numbers=left
]
TEMPORAL_CAPTION_GENERATOR = {
    "1. TEMPORAL SETTINGS": {
        "Time Intervals": [
            "Before-After",
            "Sequential Steps",
            "Time of Day",
        ],
        
        "Change Types": {
            "Natural Processes": {
                "changes": [
                    "Growth/Decay",
                    "Weather Changes",
                    "Erosion"
                ],
                "root_causes": [
                    "Biological mechanisms",
                    "Atmospheric conditions",
                    "Physical weathering"
                ]
            },
            
            "Human Activities": {
                "changes": [
                    "Construction/Demolition",
                    "Cooking/Food Preparation",
                    "Cleaning/Maintenance"
                ],
                "root_causes": [
                    "Economic needs",
                    "Survival needs",
                    "Hygiene requirements"
                ]
            },
            
            "Object Transformations": {
                "changes": [
                    "State Changes",
                    "Assembly/Disassembly",
                    "Wear and Tear"
                ],
                "root_causes": [
                    "Energy transfer",
                    "Design requirements",
                    "Friction and use"
                ]
            },
            
            "Environmental Changes": {
                "changes": [
                    "Climate Change",
                    "Habitat Modification",
                    "Pollution Effects",
                    "Ecosystem Changes"
                ],
                "root_causes": [
                    "Greenhouse gas emissions",
                    "Human expansion",
                    "Industrial activities",
                    "Species interactions"
                ]
            },
            
            "Temporal Changes": {
                "changes": [
                    "Aging Process",
                    "Historical Evolution",
                    "Day/Night Cycles",
                    "Seasonal Transitions"
                ],
                "root_causes": [
                    "Biological aging",
                    "Cumulative events",
                    "Earth's rotation",
                    "Planetary orbit"

                ]
            }
        }
    }
}
\end{lstlisting}

\paragraph{Spatial Transformations.} 
\label{app:spatial_seed}
Spatial seeds capture changes in object positions, distributions, or geometric configurations (e.g., translation, clustering, shape morphing). Each spatial change is paired with a physical or organizational root cause (e.g., ``rotation'' $\rightarrow$ ``torque effects''). 

\noindent \textbf{Spatial Transformation Generator Seeds}
\begin{lstlisting}[
    language=json,
    frame=single,
    basicstyle=\ttfamily\footnotesize,
    breaklines=true,
    numbers=left
]
SPATIAL_CAPTION_GENERATOR = {
    "1. SPATIAL SETTINGS": {
        "Spatial Dimensions": [
            "2D Planar",
            "3D Volume",
            "Multi-layered"
        ],
        
        "Change Types": {
            "Position Changes": {
                "changes": [
                    "Translation",
                    "Rotation",
                    "Reflection"
                ],
                "root_causes": [
                    "Force application",
                    "Torque effects",
                    "Symmetry operations"
                ]
            },
            
            "Spatial Distribution": {
                "changes": [
                    "Clustering",
                    "Dispersion",
                    "Alignment"
                ],
                "root_causes": [
                    "Attraction forces",
                    "Repulsion forces",
                    "Directional fields"
                ]
            },
            
            "Geometric Transformations": {
                "changes": [
                    "Shape Morphing",
                    "Angular Changes",
                    "Proportional Scaling",
                ],
                "root_causes": [
                    "Shape stress",
                    "Rotational force",
                    "Uniform pressure"
                ]
            },
        }
    }
}
\end{lstlisting}

\paragraph{Emotional Transformations.}  
\label{app:emotional_seed}
Emotional seeds encode positive and negative transitions for humans, animals, and social groups (e.g., ``fear $\rightarrow$ confidence'' caused by reassurance, or ``trust $\rightarrow$ fear'' caused by betrayal). Each transition is linked to explicit transformation triggers, enabling grounded reasoning in synthetic data.  

\noindent \textbf{Emotional Transformation Generator Seeds}
\begin{lstlisting}[
    language=json,
    frame=single,
    basicstyle=\ttfamily\footnotesize,
    breaklines=true,
    numbers=left
]
EMOTION_GENERATOR = {
    "1. MENTAL SETTINGS": {
        "Time Intervals": [
            "Immediate-Delayed",
            "Sequential Processing",
            "Short-Term",
            "Long-Term",
            "Gradual Transition",
            "Sudden Change"
        ],
        
        "Change Types": {
            "Human Emotions": {
                "positive_changes": [
                    "Sadness to Happiness",
                    "Worry to Relief",
                    "Anger to Calmness"
                ],
                "negative_changes": [
                    "Happiness to Sadness",
                    "Relief to Worry",
                    "Calmness to Anger"
                ],
                "positive_root_causes": [
                    "receiving unexpected support",
                    "finding lost items",
                    "reuniting with loved ones"
                ],
                "negative_root_causes": [
                    "experiencing loss",
                    "facing rejection",
                    "receiving bad news"
                ]
            },
            
            "Social Emotions": {
                "positive_changes": [
                    "Loneliness to Connection",
                    "Rejection to Acceptance",
                    "Isolation to Belonging",
                ],
                "negative_changes": [
                    "Connection to Loneliness",
                    "Acceptance to Rejection",
                    "Belonging to Isolation"
                ],
                "positive_root_causes": [
                    "social support",
                    "community acceptance",
                    "friendship formation"
                ],
                "negative_root_causes": [
                    "social rejection",
                    "betrayal of trust",
                    "group exclusion"
                ]
            }
        }
    }
}
\end{lstlisting}

\subsection{Ground Truth Response Quality Inspection}
\label{app:human_verification}

To better understand potential sources of label ambiguity in SD-MAR-TSE, we perform a qualitative inspection of 300 randomly sampled examples. The goal of this inspection is to identify common patterns where the provided ground-truth answer may be ambiguous or insufficiently distinguished from alternative choices.

During inspection, we observe that most samples provide a plausible explanation of the visual transformation between the paired images. However, a subset of examples exhibits ambiguity due to multiple reasonable interpretations of the observed change. This phenomenon occurs most frequently in transformations involving semantic or affective attributes, where visual evidence may support more than one textual explanation. In addition, some examples contain distractor options that are not sufficiently separated from the ground truth, leading to cases where multiple answers appear plausible.

Overall, approximately one-third of the inspected samples exhibit some level of ambiguity in answer interpretation or distractor separation. We note that such ambiguity is common in synthetic reasoning datasets \citep{Riaz_2025} that require semantic interpretation of visual changes. Importantly, this ambiguity reflects the inherent difficulty of transformation-based multimodal reasoning rather than systematic annotation failure.

Despite this noise, the dataset still provides substantial supervision signal for structured reasoning. As demonstrated in Section~\ref{sec:exp_results}, models fine-tuned on SD-MAR consistently improve multimodal analytical reasoning performance across multiple benchmarks. We therefore view this inspection as a preliminary quality analysis that helps characterize task difficulty. Future work will further improve answer generation and filtering mechanisms to reduce ambiguity and strengthen ground-truth reliability.


\paragraph{Factuality Verification and Diversity Analysis}
\label{app:fact_diversity}

To ensure the quality of data generated by the SD-MAR SDG module, we conduct both automatic filtering and qualitative inspection. For SD-MAR-TSE, we first apply CLIPScore-based filtering~\citep{hessel2021clipscore} to remove incoherent image-text pairs. We further perform a qualitative inspection of sampled examples (Appendix~\ref{app:human_verification}) to examine semantic consistency and potential ambiguity in the generated reasoning tasks. 

For SD-MAR-Math, we adopt a backtracking construction strategy grounded in real-world data sources, ensuring factual correctness through deterministic computation. Full verification details are provided in Appendix~\ref{app:dataset_verification}. To assess overall data diversity, we analyze the distribution of change types in SD-MAR-TSE and categorize SD-MAR-Math samples by reasoning complexity levels. Detailed diversity statistics are reported in Appendix~\ref{app:MAR_diversity}.

\subsection{Dataset Filter and Alignment Verification}
\label{app:dataset_verification}

To ensure semantic coherence between captions and images, we applied CLIP-based filtering with a threshold of 15. Table~\ref{tab:clip_filter_summary} summarizes the statistics for each transformation domain.

\subsubsection{Temporal}
\begin{itemize}
    \item Original size: 2,074 pairs
    \item Filtered size: 2,068 pairs (6 removed)
\end{itemize}
\textbf{Stage 1 CLIP scores:} Avg.~21.38, Min.~15.11 (pair 1073), Max.~28.03 (pair 843).  
\textbf{Stage 2 CLIP scores:} Avg.~21.86, Min.~15.63 (pair 705), Max.~29.00 (pair 1772).  

\subsubsection{Spatial}
\begin{itemize}
    \item Original size: 2,990 pairs
    \item Filtered size: 2,990 pairs (0 removed)
\end{itemize}
\textbf{Stage 1 CLIP scores:} Avg.~24.84, Min.~16.77 (pair 1221), Max.~30.92 (pair 2137).  
\textbf{Stage 2 CLIP scores:} Avg.~25.01, Min.~16.27 (pair 1177), Max.~31.25 (pair 329).  

\subsubsection{Emotion}
\begin{itemize}
    \item Original size: 937 pairs
    \item Filtered size: 937 pairs (0 removed)
\end{itemize}
\textbf{Stage 1 CLIP scores:} Avg.~22.60, Min.~17.13 (pair 705), Max.~27.18 (pair 225).  
\textbf{Stage 2 CLIP scores:} Avg.~22.30, Min.~17.22 (pair 459), Max.~27.03 (pair 391).  

\begin{table}[h!]
\centering
\small
\begin{tabular}{l|c|c|c|c|c}
\toprule
Domain & Orig. & Final & Removed & Stage 1 (A/Mi/Ma) & Stage 2 (A/Mi/Ma) \\
\midrule
Temporal & 2074 & 2068 & 6 & 21.38 / 15.11 / 28.03 & 21.86 / 15.63 / 29.00 \\
Spatial  & 2990 & 2990 & 0 & 24.84 / 16.77 / 30.92 & 25.01 / 16.27 / 31.25 \\
Emotion  & 937  & 937  & 0 & 22.60 / 17.13 / 27.18 & 22.30 / 17.22 / 27.03 \\
\bottomrule
\end{tabular}
\caption{CLIP-based filtering statistics across transformation domains. Stage~1 measures similarity between the base image and the generated image, while Stage~2 measures similarity between the edited image and the target prompt. A/Mi/Ma denote Avg/Min/Max.}
\label{tab:clip_filter_summary}
\end{table}

\subsection{SD-MAR Diversity Analysis}
\label{app:MAR_diversity}

To validate the diversity and coverage of our synthetic dataset, we report statistics across temporal, spatial, emotional, and mathematical domains.

\subsubsection{Temporal Diversity}
We categorize temporal reasoning into six settings: \textit{Before--After}, \textit{Time of Day}, \textit{Historical Periods}, \textit{Sequential Steps}, \textit{Seasons}, and \textit{Life Stages}. Each setting includes six transformation change types. Overall distribution is balanced, with \textit{Temporal Changes} (387), \textit{Environmental Changes} (358), and \textit{Social and Cultural Changes} (355) as the largest groups.  
Breakdowns show complementary emphases: \textit{Before--After} focuses on natural processes and human activities, while \textit{Historical Periods} highlight social and cultural transformations. \textit{Seasons} and \textit{Life Stages} diversify coverage by mixing physical and social processes. The aggregated distribution is shown in Fig.~\ref{fig:temporal_spatial_causality}.

\subsubsection{Spatial Diversity}
\begin{figure}[h]
    \centering
    \includegraphics[width=0.95\linewidth]{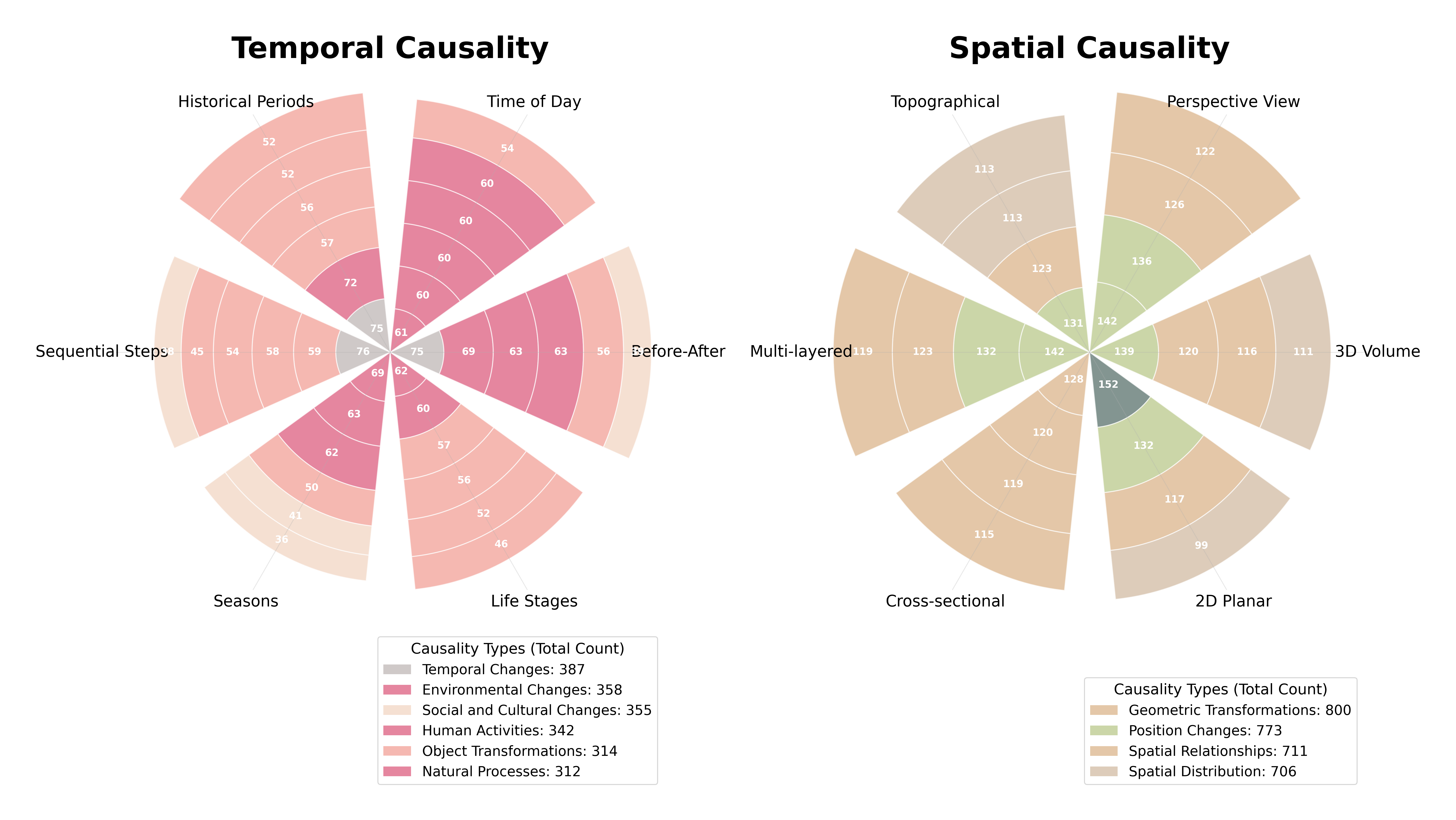}
    \caption{Temporal and spatial transformation distribution across categories and settings.}
    \label{fig:temporal_spatial_causality}
\end{figure}

Spatial reasoning spans four categories: \textit{Geometric Transformations} (800), \textit{Position Changes} (773), \textit{Spatial Relationships} (711), and \textit{Spatial Distribution} (706).  
Coverage is evenly distributed across six spatial views (2D Planar, 3D Volume, Multi-layered, Cross-sectional, Topographical, Perspective). For example, \textit{Geometric Transformations} dominate in 2D Planar (30.4\%), while \textit{Position Changes} are most frequent in Perspective View (27.0\%). Together with the temporal breakdown, these distributions ensure exposure to both geometric and relational reasoning constraints (Fig.~\ref{fig:temporal_spatial_causality}).

\subsubsection{Emotion Diversity}
\begin{figure}[h]
    \centering
    \includegraphics[width=0.95\linewidth]{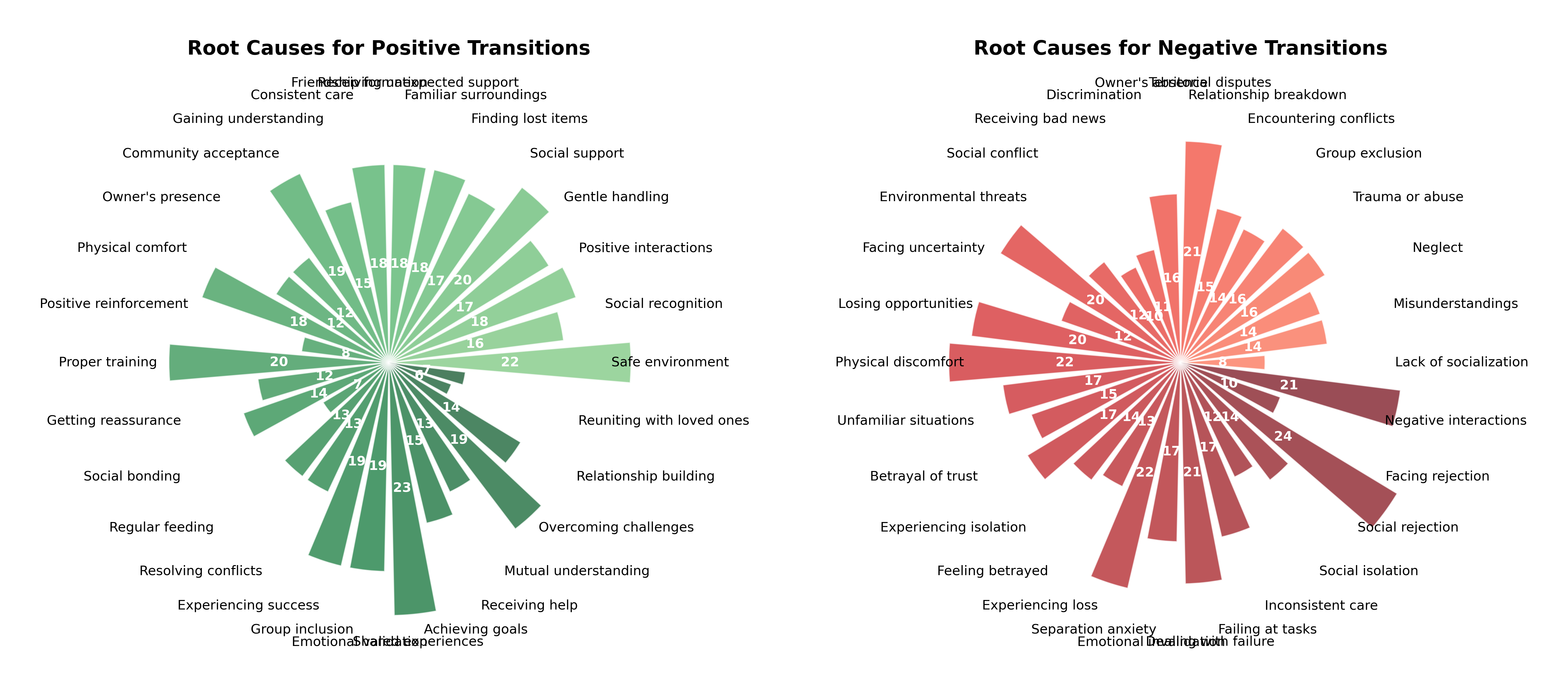}
    \caption{Analytical factor of emotional transitions: (left) positive, (right) negative.}
    \label{fig:conditional_emotion_analysis}
\end{figure}

Emotional transitions are nearly balanced between positive (462; 49.3\%) and negative (475; 50.7\%), preventing sentiment polarity bias. Positive transitions include, e.g., \textit{``sadness to happiness''}, while negative transitions cover inverses such as \textit{``confidence to fear''}.  
Root causes---such as \textit{receiving support} for positive and \textit{social rejection} for negative---span social, personal, and environmental contexts. Fig.~\ref{fig:emotion_causality_math} shows the overall polarity distribution, while Fig.~\ref{fig:conditional_emotion_analysis} further decomposes root causes for both positive and negative transitions.

\subsubsection{Math Difficulty Distribution}
\begin{figure}[h]
    \centering
    \includegraphics[width=0.95\linewidth]{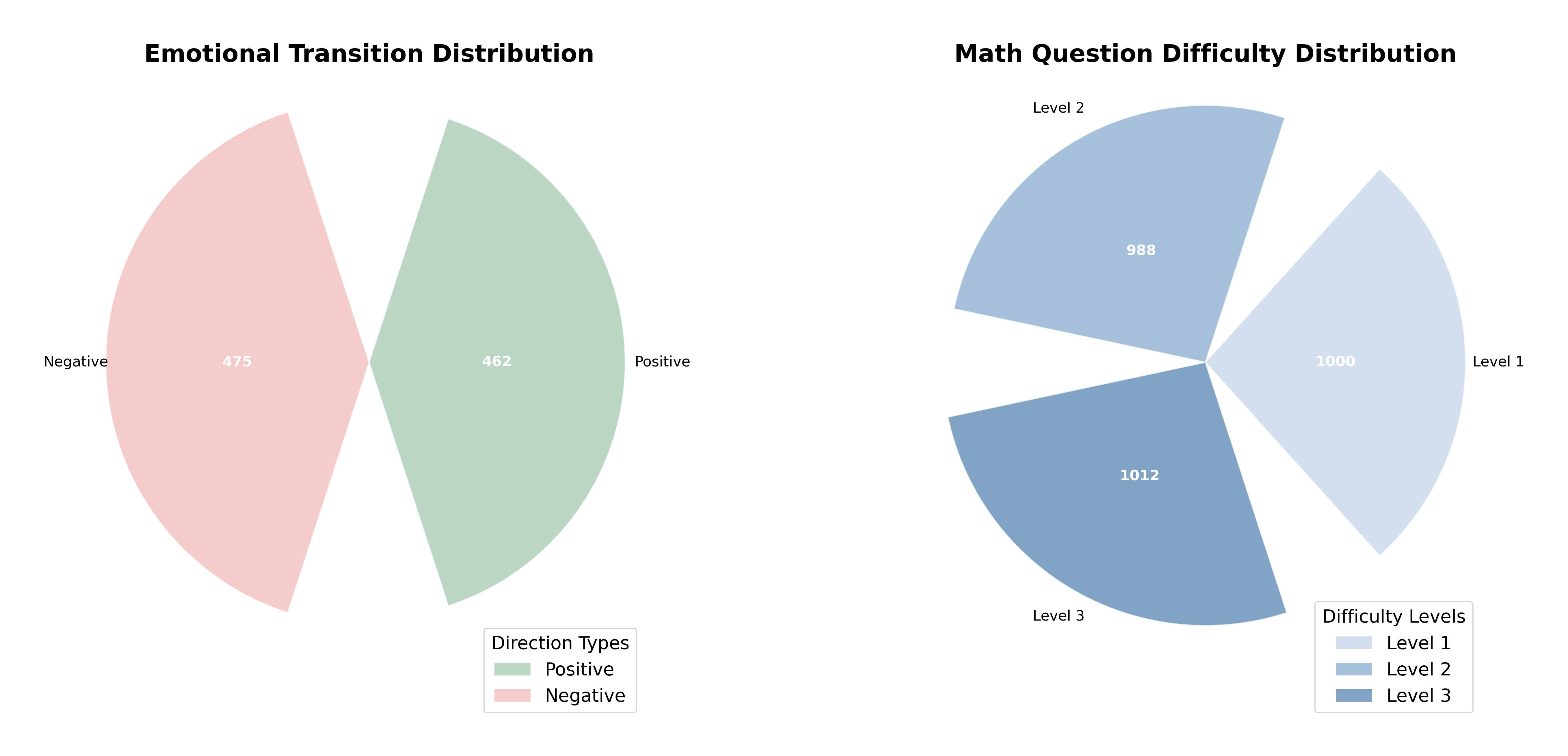}
    \caption{Emotion transformation distribution (positive vs.\ negative) and math difficulty distribution (Level 1--3).}
    \label{fig:emotion_causality_math}
\end{figure}

The SD-MAR-Math dataset is stratified into three difficulty levels with equal size: Level~1 (absolute difference), Level~2 (percentage change), and Level~3 (multi-country growth rate). Each contains 5{,}000 questions (33.3\% each), ensuring balanced task complexity and controlled error compounding. The distribution is illustrated in Fig.~\ref{fig:emotion_causality_math}.


\subsubsection{Option Analysis}
We further analyze the token distribution of multiple-choice options to verify consistency between correct and incorrect answers. 
The average token counts are nearly identical across categories: all options (2.03), correct options (2.01), and incorrect options (2.03). This indicates that option length does not introduce bias or spurious cues for model training.  

In total, we analyzed 21,580 options, consisting of 5,395 correct and 16,185 incorrect answers, confirming that the dataset maintains a balanced and unbiased structure in terms of token length across different option types.

\clearpage
\subsection{Dataset Samples}
\label{app:dataset_samples}
\subsubsection{Temporal Transformation}
\label{app:temporal_samples}

\newenvironment{questionbox}[1]{%
    \begin{tcolorbox}[
        title=#1,
        colback=white,
        colframe=blue!75!black,
        fonttitle=\bfseries
    ]
}{%
    \end{tcolorbox}
}

\newenvironment{choicelist}{%
    \begin{enumerate}[label=(\Alph*)]
}{%
    \end{enumerate}
}

\begin{figure}[htb]
    \centering
    
    \begin{minipage}{0.9\linewidth}
        \centering
        \includegraphics[width=\linewidth]{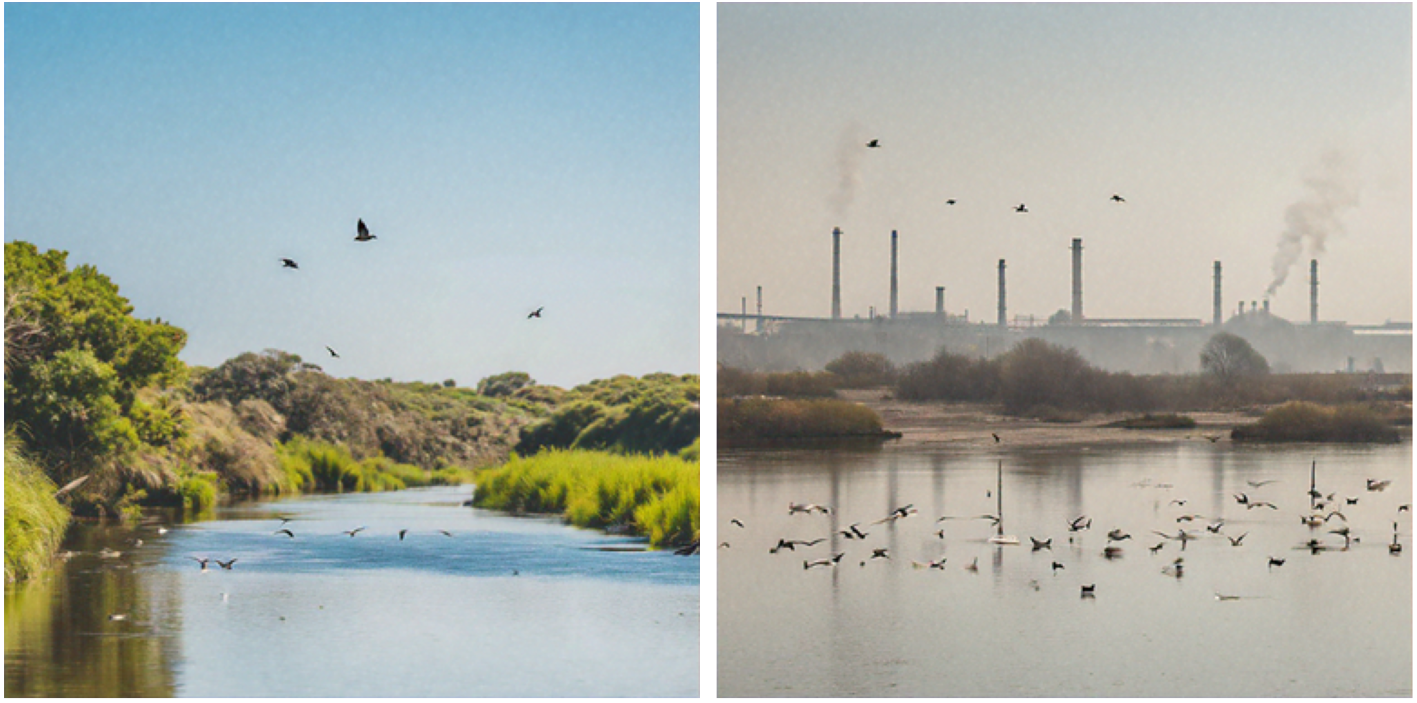}
        
        \vspace{0.5em}
        
        \begin{questionbox}{Temporal Transformation Question Sample}
            What is the primary cause of the changes observed between these two states?
            
            \begin{choicelist}
                \item Molecular interactions
                \item Market forces
                \item Industrial activities
                \item Atmospheric conditions
            \end{choicelist}
            
            \textbf{Correct Answer:} C (Industrial activities)
        \end{questionbox}
    \end{minipage}
    
    \caption{Temporal Transformation}
    \label{fig:temporal_causality_1}
\end{figure}

\paragraph{Reasoning Trace}

\textbf{Observation Analysis}
\begin{itemize}
    \item \textit{Initial State:} A pristine natural landscape with clear blue skies, clean flowing water in a river, and abundant wildlife. Fish swim in the transparent water while birds soar overhead. The surrounding vegetation is lush and healthy, with diverse plant species thriving in the unpolluted environment.
    
    \item \textit{Final State:} The same landscape is severely degraded by industrial pollution. The sky is hazy with smog, and the river has turned murky due to chemical runoff and waste. Dead fish float on the water's surface, birds are absent, and vegetation has withered due to airborne toxins and contaminated soil. Factory smokestacks continuously emit dark smoke into the atmosphere.
    
    \item \textit{Key Differences:} The scene transitions from a clean and biodiverse ecosystem to a polluted and degraded environment, characterized by water contamination, loss of wildlife, and vegetation decline.
\end{itemize}

\textbf{Causal Analysis}
\begin{itemize}
    \item \textit{Identified Changes:} The transformation from a pristine natural state to a polluted and degraded environment.
    
    \item \textit{Primary Cause Reasoning:} Industrial activities drive these changes by introducing pollutants into air, water, and soil, leading to ecosystem degradation.
\end{itemize}

\begin{figure}[htb]
    \centering
    
    \begin{minipage}{0.9\linewidth}
        \centering
        \includegraphics[width=\linewidth]{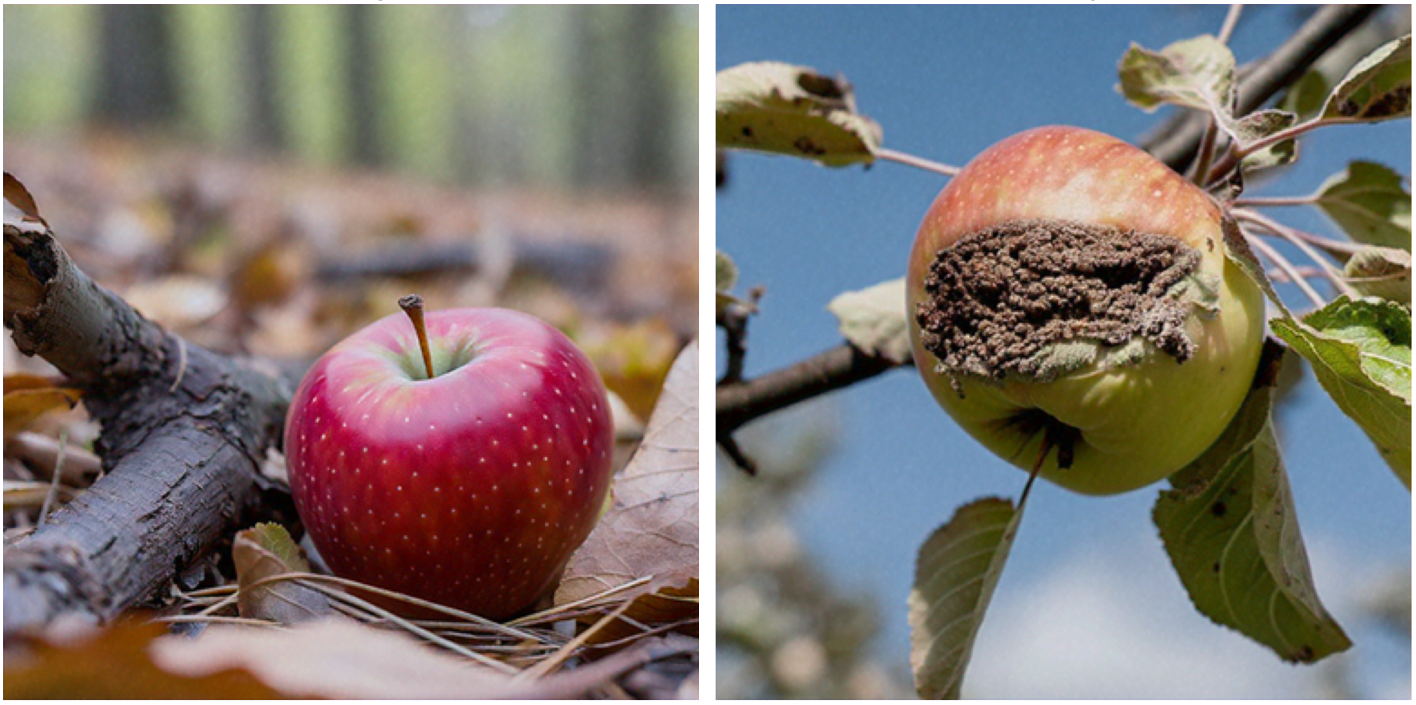}
        
        \vspace{0.5em}
        
        \begin{questionbox}{Temporal Transformation Question Sample}
            What is the primary cause of the changes observed between these two states?
            
            \begin{choicelist}
                \item Microbial activity
                \item Chemical exposure
                \item Molecular interactions
                \item Population growth
            \end{choicelist}
            
            \textbf{Correct Answer:} A (Microbial activity)
        \end{questionbox}
    \end{minipage}
    
    \caption{Temporal Transformation}
    \label{fig:temporal_causality}
\end{figure}

\paragraph{Reasoning Trace}

\textbf{Observation Analysis}
\begin{itemize}
    \item \textit{Initial State:} A fresh apple with vibrant red skin and firm flesh lies on the forest floor after recently falling from a tree. Its surface is intact, smooth, and glossy, with no visible signs of decay.
    
    \item \textit{Final State:} The same apple exhibits advanced decomposition due to microbial activity. The skin has darkened and wrinkled, with patches of fuzzy mold growth. The flesh has softened and partially collapsed as bacteria and fungi break down the cellular structure.
    
    \item \textit{Key Differences:} The apple transitions from a fresh, intact state to a decomposed state characterized by discoloration, mold growth, and structural degradation.
\end{itemize}

\textbf{Causal Analysis}
\begin{itemize}
    \item \textit{Identified Changes:} The transformation from a fresh apple to a decomposed state.
    
    \item \textit{Primary Cause Reasoning:} Microbial activity drives these changes by decomposing organic material, breaking down cellular structures, and facilitating nutrient recycling.
\end{itemize}

\clearpage
\subsubsection{Spatial Transformation}
\label{app:spatial_samples}

    
    

\begin{figure}[h!]
    \centering
    
    \begin{minipage}{0.9\linewidth}
        \centering
        \includegraphics[width=\linewidth]{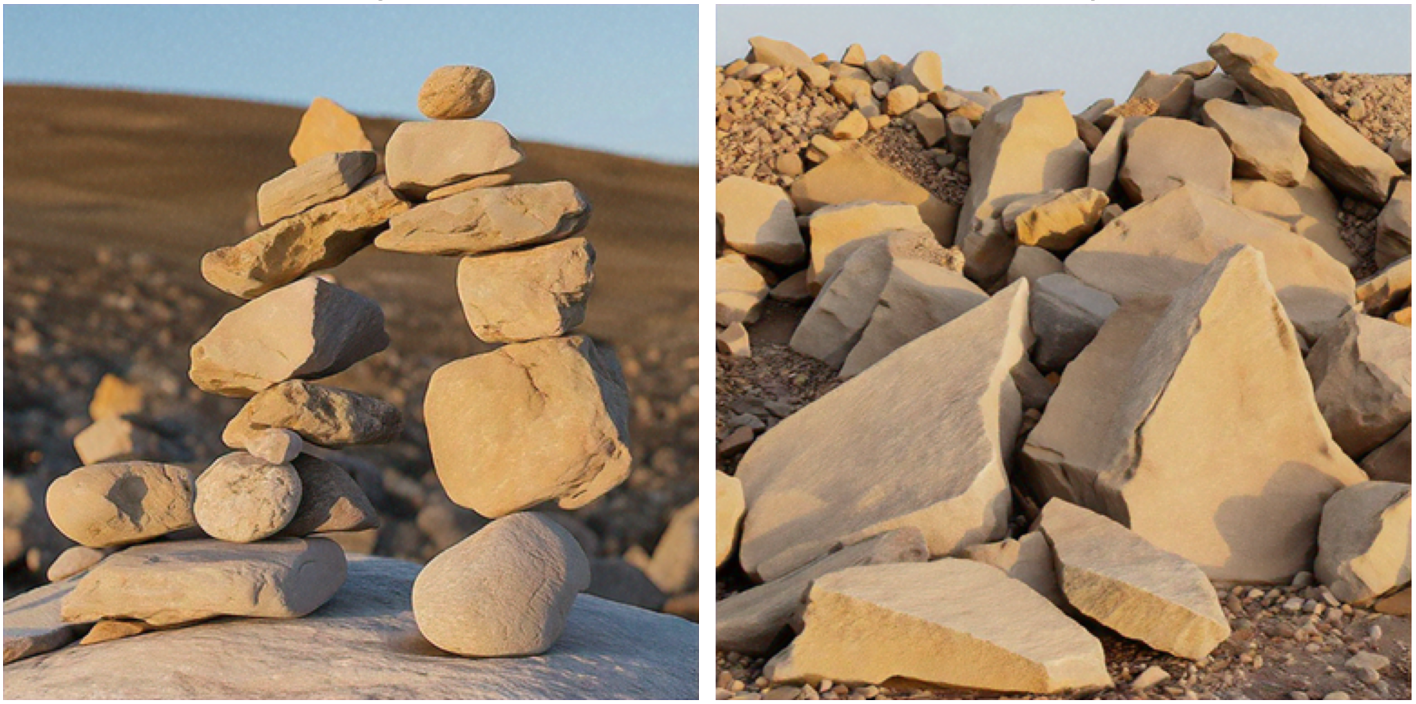}
        
        \vspace{0.5em}
        \hrule
        \vspace{0.5em}
        
        \begin{questionbox}{Spatial Transformation Question}
            What is the primary cause of the changes observed between these two states?
            
            \begin{choicelist}
                \item Torque effects
                \item Equilibrium forces
                \item Organizational rules
                \item Symmetry operations
            \end{choicelist}
            
            \textbf{Correct Answer:} B (Equilibrium forces)
        \end{questionbox}
    \end{minipage}
    
    \caption{Spatial Transformation}
    \label{fig:spatial_causality}
\end{figure}

\paragraph{Reasoning Trace}

\textbf{Observation Analysis}
\begin{itemize}
    \item \textit{Initial State:} Several rocks of varying sizes are precariously stacked in an unstable formation, with the center of mass significantly shifted to one side. The arrangement appears temporarily balanced but unstable.
    
    \item \textit{Final State:} The rocks have collapsed into a more stable configuration with a wider base and a lower center of mass. Larger rocks form the foundation, while smaller rocks settle on top, maximizing stability through increased contact support.
    
    \item \textit{Key Differences:} The system transitions from an unstable, imbalanced stacking to a stable, grounded configuration characterized by a lower center of mass and improved structural support.
\end{itemize}

\textbf{Causal Analysis}
\begin{itemize}
    \item \textit{Identified Changes:} The rearrangement of rocks from an unstable configuration to a stable equilibrium state.
    
    \item \textit{Primary Cause Reasoning:} Equilibrium forces drive this transformation by redistributing mass and contact forces, leading the system toward a lower-energy, stable configuration.
\end{itemize}

\clearpage
\begin{figure}[h!]
    \centering
    
    \begin{minipage}{0.9\linewidth}
        \centering
        \includegraphics[width=\linewidth]{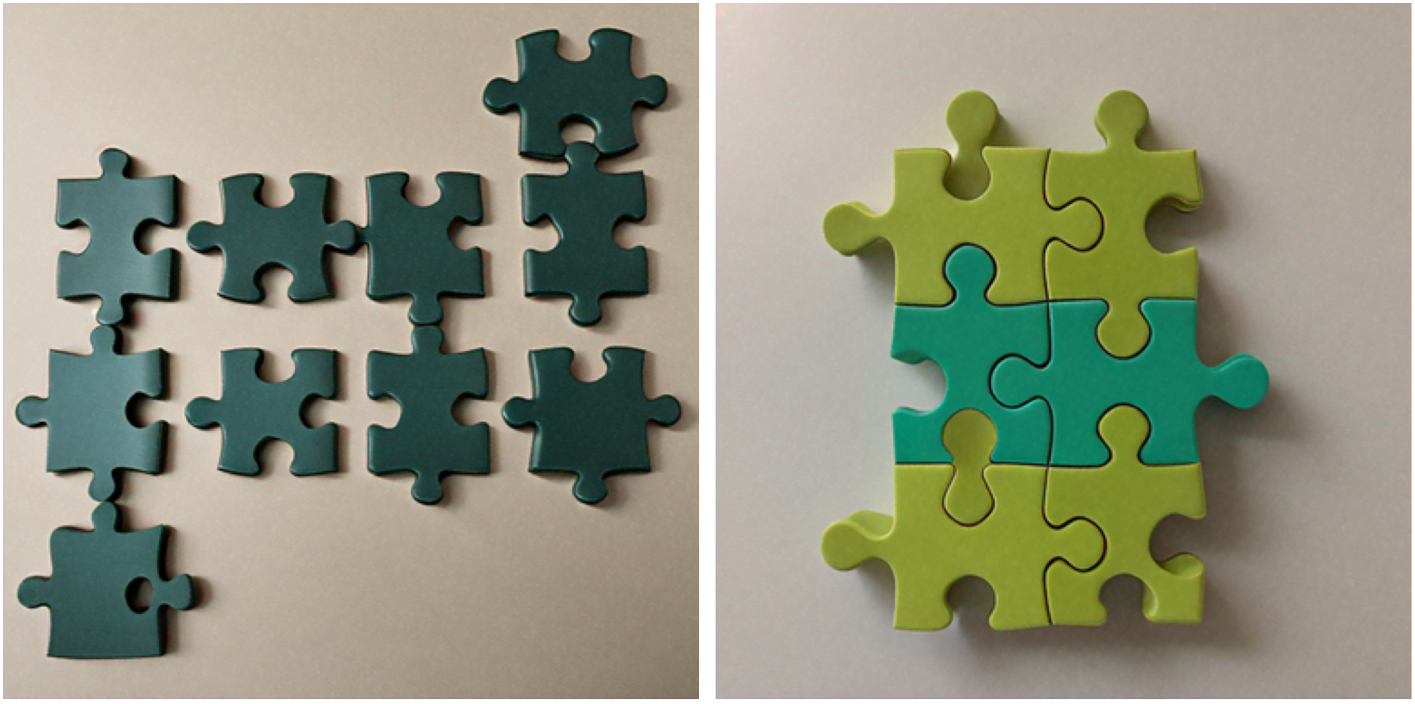}
        
        \vspace{0.5em}
        \hrule
        \vspace{0.5em}
        
        \begin{questionbox}{Spatial Configuration Analysis}
            What is the primary cause of the changes observed between these two states?
            
            \begin{choicelist}
                \item Force application
                \item Component fitting
                \item Repulsion forces
                \item Connectivity requirements
            \end{choicelist}
            
            \textbf{Correct Answer:} B (Component fitting)
        \end{questionbox}
    \end{minipage}
    
    \caption{Component Assembly Analysis}
    \label{fig:spatial_component_fitting}
\end{figure}

\paragraph{Reasoning Trace}

\textbf{Observation Analysis}
\begin{itemize}
    \item \textit{Initial State:} Several distinct puzzle pieces with complementary protrusions and indentations are arranged separately on a surface. Each piece has a unique shape with tabs and slots designed for interlocking.
    
    \item \textit{Final State:} The puzzle pieces are assembled into a cohesive multi-layered structure. Tabs are inserted into corresponding slots, forming a stable interlocked configuration where pieces support each other in three dimensions.
    
    \item \textit{Key Differences:} The system transitions from a set of disjoint components to an integrated, interlocked structure characterized by precise alignment and structural stability.
\end{itemize}

\textbf{Causal Analysis}
\begin{itemize}
    \item \textit{Identified Changes:} The assembly of separate components into a stable, interlocking configuration.
    
    \item \textit{Primary Cause Reasoning:} Component fitting drives this transformation by aligning compatible shapes and enabling interlocking connections that ensure structural stability.
\end{itemize}

\clearpage
\subsubsection{Emotion Transformation}
\label{app:emotion_samples}

\begin{figure}[htb]
    \centering
    
    \begin{minipage}{0.9\linewidth}
        \centering
        \includegraphics[width=\linewidth]{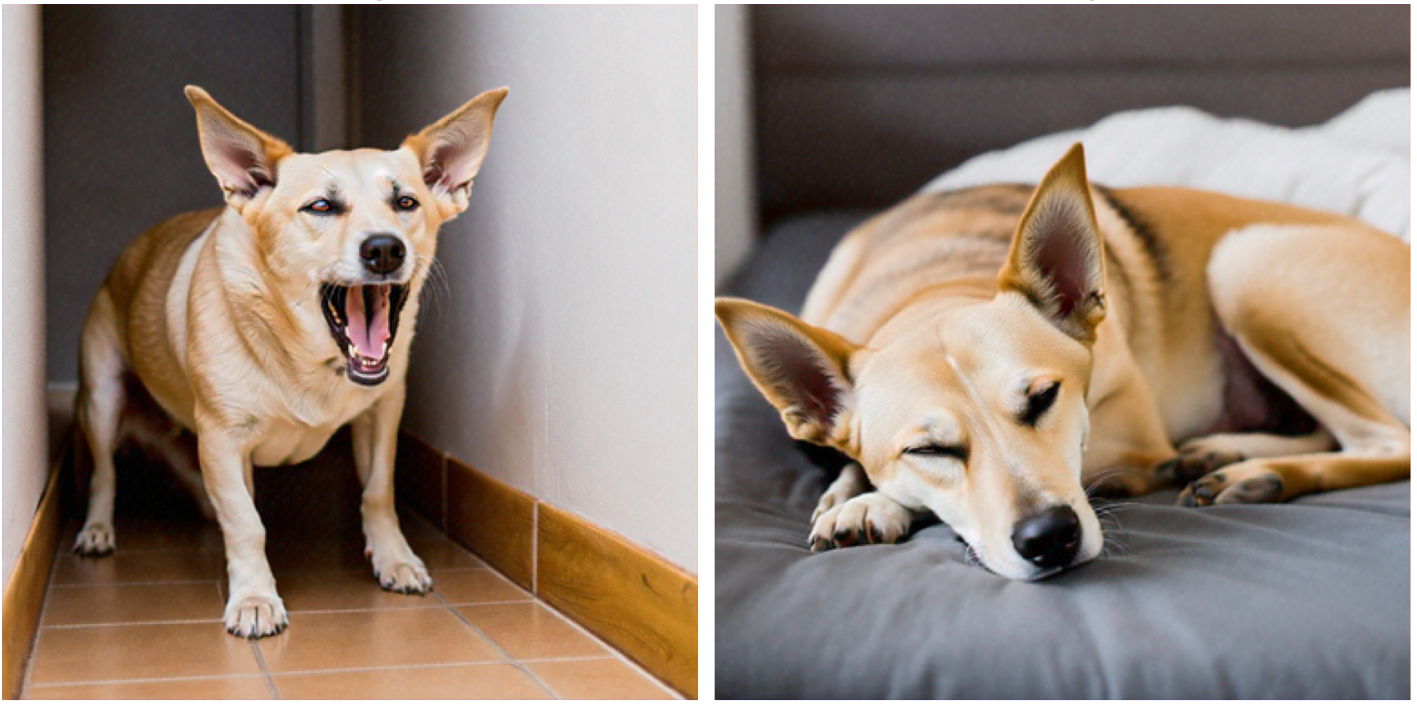}
        \vspace{0.5em}
        \hrule
        \vspace{0.5em}
        
        \begin{questionbox}{Emotional Response Analysis}
            What is the primary cause of the changes observed between these two states?
            
            \begin{choicelist}
                \item Safe environment
                \item Overcoming challenges
                \item Community acceptance
                \item Relationship building
            \end{choicelist}
            
            \textbf{Correct Answer:} A (Safe environment)
        \end{questionbox}
    \end{minipage}
    
    \caption{Emotional Response Analysis}
    \label{fig:emotional_safety}
\end{figure}

\paragraph{Reasoning Trace}

\textbf{Observation Analysis}
\begin{itemize}
    \item \textit{Initial State:} A trembling dog with pinned-back ears and wide, anxious eyes, cowering in a corner with visible body tension and heavy panting in response to perceived threats.
    
    \item \textit{Final State:} The same dog is relaxed with a soft expression, ears in a natural position, and body loosely curled on a comfortable bed, breathing steadily with half-closed eyes in a calm environment.
    
    \item \textit{Key Differences:} The dog transitions from a fearful, stress-induced state to a relaxed and secure state, characterized by reduced tension, calmer posture, and stable breathing.
\end{itemize}

\textbf{Causal Analysis}
\begin{itemize}
    \item \textit{Identified Changes:} The shift from a high-stress defensive state to a calm and relaxed condition.
    
    \item \textit{Primary Cause Reasoning:} A safe environment enables this transformation by reducing perceived threats, allowing the dog to relax and return to a stable behavioral state.
\end{itemize}

\clearpage
\subsubsection{Math Comparison Analysis}
\label{app:math_samples_trace}

    
    

\begin{figure}[htb]
    \centering
    
    \begin{minipage}{0.9\linewidth}
        \centering
        \includegraphics[width=\linewidth]{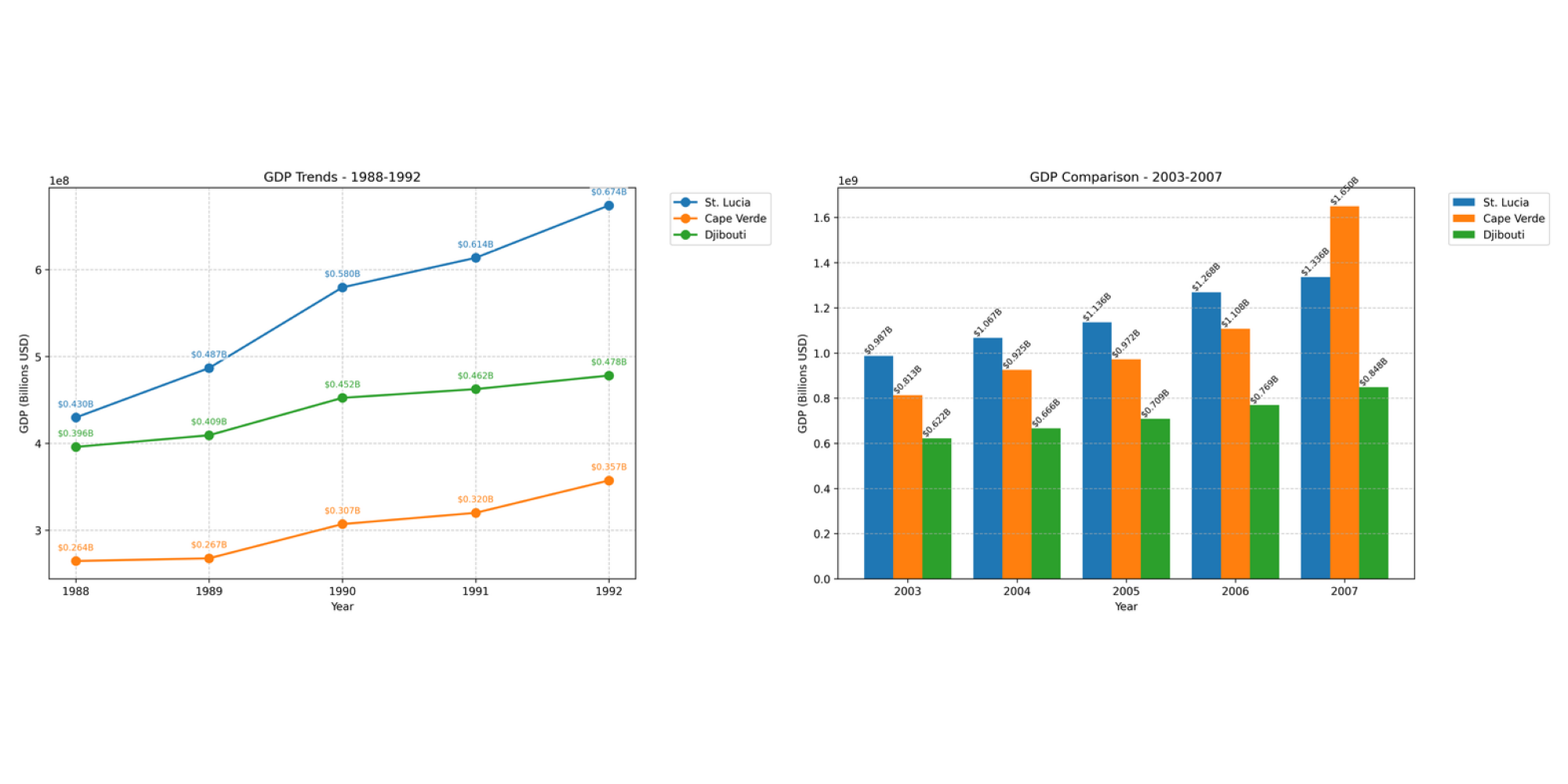}
        
        \vspace{0.5em}
        \hrule
        \vspace{0.5em}
        
        \begin{questionbox}{Economic Growth Analysis}
            Which country showed the strongest GDP growth between 1988-1992 and 2003-2007?
            
            \begin{choicelist}
                \item St. Lucia with 108.141\% growth
                \item Cape Verde with 260.749\% growth
                \item Djibouti with 64.415\% growth
                \item Cape Verde with 273.165\% growth
            \end{choicelist}
            
            \textbf{Correct Answer:} B (Cape Verde with 260.749\% growth)
        \end{questionbox}
    \end{minipage}
    
    \caption{Economic Growth Analysis}
    \label{fig:gdp_growth_analysis}
\end{figure}

\paragraph{Ground-Truth Analysis}

\textbf{1. GDP Analysis for All Countries}

\textbf{St. Lucia}
\begin{itemize}
    \item \textit{1988--1992:}
    \begin{itemize}
        \item 1988: \$0.430 billion
        \item 1989: \$0.487 billion
        \item 1990: \$0.580 billion
        \item 1991: \$0.614 billion
        \item 1992: \$0.674 billion
    \end{itemize}
    Average: \$0.557 billion

    \item \textit{2003--2007:}
    \begin{itemize}
        \item 2003: \$0.987 billion
        \item 2004: \$1.067 billion
        \item 2005: \$1.136 billion
        \item 2006: \$1.268 billion
        \item 2007: \$1.336 billion
    \end{itemize}
    Average: \$1.159 billion

    \item \textbf{Growth rate:} 108.141\%
\end{itemize}

\textbf{Cape Verde}
\begin{itemize}
    \item \textit{1988--1992:}
    \begin{itemize}
        \item 1988: \$0.264 billion
        \item 1989: \$0.267 billion
        \item 1990: \$0.307 billion
        \item 1991: \$0.320 billion
        \item 1992: \$0.357 billion
    \end{itemize}
    Average: \$0.303 billion

    \item \textit{2003--2007:}
    \begin{itemize}
        \item 2003: \$0.813 billion
        \item 2004: \$0.925 billion
        \item 2005: \$0.972 billion
        \item 2006: \$1.108 billion
        \item 2007: \$1.650 billion
    \end{itemize}
    Average: \$1.094 billion

    \item \textbf{Growth rate:} 260.749\%
\end{itemize}

\textbf{Djibouti}
\begin{itemize}
    \item \textit{1988--1992:}
    \begin{itemize}
        \item 1988: \$0.396 billion
        \item 1989: \$0.409 billion
        \item 1990: \$0.452 billion
        \item 1991: \$0.462 billion
        \item 1992: \$0.478 billion
    \end{itemize}
    Average: \$0.440 billion

    \item \textit{2003--2007:}
    \begin{itemize}
        \item 2003: \$0.622 billion
        \item 2004: \$0.666 billion
        \item 2005: \$0.709 billion
        \item 2006: \$0.769 billion
        \item 2007: \$0.848 billion
    \end{itemize}
    Average: \$0.723 billion

    \item \textbf{Growth rate:} 64.415\%
\end{itemize}

\textbf{2. Conclusion}
\begin{itemize}
    \item Cape Verde showed the highest growth rate at 260.749\%
\end{itemize}

\textbf{3. Country Rankings by Growth Rate}
\begin{enumerate}
    \item Cape Verde: 260.749\%
    \item St. Lucia: 108.141\%
    \item Djibouti: 64.415\%
\end{enumerate}

\textbf{4. Visualization Details}
\begin{itemize}
    \item 1988--1992: line\_chart chart showing GDP comparison
    \item 2003--2007: bar\_chart chart showing GDP comparison
\end{itemize}

\subsection{Generation Comparison Between Base Models and MAR-Finetuned Models}
\label{app:generation_comparison}

To illustrate the effect of SD-MAR fine-tuning, we compare generations from base models (Qwen2.5-VL-7B, InternVL3-8B) with their fine-tuned counterparts, alongside the reference ground-truth (denoted as SD-MAR Groundtruth). For each evaluation sample from SD-MAR-Math and SD-MAR-TSE, we present three generations: (i) base model outputs, denoted as Qwen-Base and InternVL-Base; (ii) fine-tuned outputs, where all models are trained on SD-MAR-Mix, denoted as Mix$\rightarrow$Mix; and (iii) method-specific variants. In particular, Qwen-GRPO-lite (Mix$\rightarrow$Mix) refers to Qwen2.5-VL-7B fine-tuned with SD-MAR-Mix using GRPO-lite, while Qwen-GRPO-lite-BDA (Mix$\rightarrow$Mix) denotes the same backbone trained with GRPO-lite+BDA. The same naming convention is applied to the InternVL model family: InternVL-GRPO-lite (MIX→MIX) ,  InternVL-GRPO-lite-BDA (MIX→MIX).

\subsubsection{MAR-Math}
\label{sec:data_sample_math}

\begin{figure}[h]
    \centering
    \includegraphics[width=0.85\linewidth]{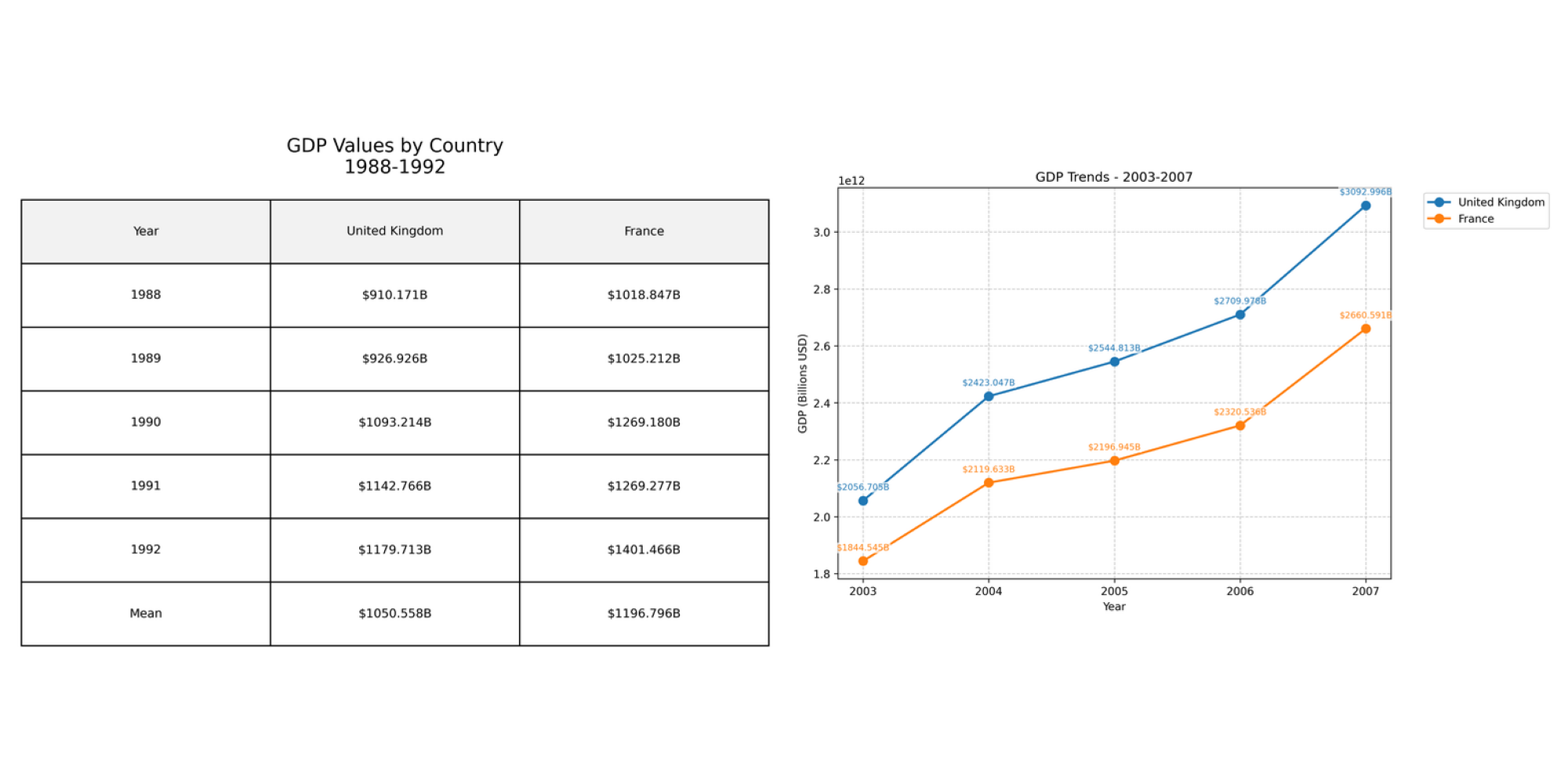}
    \caption{Example from MAR-Math}
    \label{fig:generation_example}
\end{figure}

\begin{tcolorbox}[title=Groundtruth, colframe=black!40, colback=gray!5, coltitle=black]
\textbf{MAR-Math Groundtruth}
\small
\begin{verbatim}
What was the difference in United Kingdom's GDP between the start 
of 1988-1992 and 2003-2007?

Multiple Choice Options:
(A) $1,146.534 billion
(B) $2,293.067 billion
(C) $4,012.867 billion
(D) $573.267 billion
Options:['A', 'B', 'C', 'D']
Ground Truth: 
1. GDP Values for Each Period:

United Kingdom in 1988-1992:
   - 1988: $910.171 billion
   - 1989: $926.926 billion
   - 1990: $1,093.214 billion
   - 1991: $1,142.766 billion
   - 1992: $1,179.713 billion

United Kingdom in 2003-2007:
   - 2003: $2,056.705 billion
   - 2004: $2,423.047 billion
   - 2005: $2,544.813 billion
   - 2006: $2,709.978 billion
   - 2007: $3,092.996 billion

2. Calculation:
   Initial GDP in 2003-2007: $2,056.705 billion
   Initial GDP in 1988-1992: $910.171 billion
   Absolute difference: $1,146.534 billion
4. Visualization Details:
   - 1988-1992: table chart showing GDP comparison
   - 2003-2007: line_chart chart showing GDP comparison

Answer is A
\end{verbatim}
\end{tcolorbox}

\begin{tcolorbox}[title=Qwen Responses, colframe=black!40, colback=gray!5, coltitle=black]
\textbf{Qwen-Base}
\small
\begin{verbatim}
<reasoning>
1. **Initial Observation**:
   - **Image 1 (1988-1992)**: The table shows GDP values 
     for the United Kingdom and France from 1988 to 1992.
     - For the UK, the mean GDP value is $1050.558B.
     - For France, the mean GDP value is $1196.796B.

   - **Image 2 (2003-2007)**: The line graph shows GDP 
     trends for the United Kingdom and France from 2003 
     to 2007.
     - For the UK, the GDP in 2003 is $2056.705B and in 
       2007 it is $3092.996B.
     - For France, the GDP in 2003 is $1844.545B and in 
       2007 it is $2660.591B.

2. **Analysis of Changes**:
   - The question asks for the difference in the UK's GDP 
     between the start of 1988-1992 ($1050.558B) and 
     2003-2007 ($2056.705B).

3. **Option Evaluation**:
   - **Option A**: $1,146.534B → $2056.705 - $1050.558 = $1006.147
   - **Option B**: $2,293.067B → mismatch
   - **Option C**: $4,012.867B → mismatch
   - **Option D**: $573.267B → mismatch

4. **Reflection**:
   - Calculation method correct, but wrong start year used.
   - All units consistent.

5. **Conclusion**:
   - No correct option matched. Selected D.
</reasoning>
<answer>D</answer>
\end{verbatim}
\end{tcolorbox}

\begin{tcolorbox}[colframe=black!40, colback=gray!5]
\label{box:qwen_grpolite_noBDA_math}
\textbf{Qwen-GRPO-lite(w/o)BDA (MIX→MIX)}
\small
\begin{verbatim}
<reasoning>
1. **Initial Observation:** 
   - For 1988, the GDP for the UK was approximately $910 billion.
   - For 2003, the GDP for the UK was around $2,056 billion. We need 
   to find the difference between these two values, which is roughly 
   $2,056 - $910 = $1,146.

2. **Analysis of Changes:** 
   Based on our rough calculation, we're looking for something close 
   to $1,146. Looking at the closest options, we match it with 
   option A ($1,146.534), option B ($2,293.067), option C ($4,012.867), 
   and option D ($573.267). 
   The closest match appears to be option A ($1,146.534).

3. **Option Evaluation:** 
   - Option A: $1,146.534 (too large but close)
   - Option B: $2,293.067 (too large)
   - Option C: $4,012.867 (too large)
   - Option D: $573.267 (too small)

Based on matching, option A ($1,146.534) seems closest, and matches 
with our calculated value.

4. **Reflection:** 
   Yes, the calculation method seems appropriate here as it correctly 
   captures the comparison between the two periods.

5. Based on our matching, the correct option is A.

</reasoning>
<answer>A</answer>
\end{verbatim}
\end{tcolorbox}

\begin{tcolorbox}[colframe=black!40, colback=gray!5]
\label{box:qwen_grpolite_math}
\textbf{Qwen-GRPO-lite (MIX→MIX)}
\small
\begin{verbatim}
<reasoning>
First, calculate United Kingdom's GDP at the start of 1988-1992:
- Start of 1988-1992: $910.171 billion

Next, calculate United Kingdom's GDP at the start of 2003-2007 (2003):
- Start of 2003-2007: $2056.705 billion

Difference in United Kingdom's GDP:
$2056.705 billion (2003) - $910.171 billion (1988) = $1146.534 billion

Thus, the difference in United Kingdom's GDP between the start of 
1988-1992 and 2003-2007 was closest to $1,146.534 billion.

</reasoning>
<answer>A</answer>
\end{verbatim}
\end{tcolorbox}

\begin{tcolorbox}[title=InternVL Responses, colframe=black!40, colback=gray!5, coltitle=black]
\textbf{InternVL-Base}
\small
\begin{verbatim}
<reasoning>
1. Initial Observation:
   - Initial State (1988-1992):
     - United Kingdom GDP: $1050.558B (Mean)
   - Final State (2003-2007):
     - United Kingdom GDP: $3092.996B (2007)

2. Analysis of Changes:
   - The United Kingdom's GDP increased from $1050.558B in 1988-1992 
   to $3392.996B in 2007.

3. Option Evaluation:
   - Option A: $1,146.534 billion
     - Incorrect. The difference is much larger.
   - Option B: $2,293.067 billion
     - Incorrect. The difference is still not large enough.
   - Option C: $4,002.867 billion
     - Incorrect. The difference is even larger than the actual 
     increase.
   - Option D: $573.267 billion
     - Correct. The difference is $3392.996B - $1050.558B = $2342.438B, 
     which is closest to $2,293.067 billion.

4. Reflection:
   - The calculation method is appropriate for this problem.
   - Units are consistent and correct.
   - The result makes logical sense given the context.
   - All relevant factors have been considered.

5. Conclusion:
   - The correct answer is Option D.
</reasoning>
<answer>D</answer>

\end{verbatim}
\end{tcolorbox}

\begin{tcolorbox}[colframe=black!40, colback=gray!5]
\textbf{InternVL-GRPO-lite(w/o) BDA (MIX→MIX)}
\small
\begin{verbatim}
<reasoning>
1. Initial Observation:
   - Initial State (1988-1992):
     - 1988: United Kingdom GDP = $910.171 billion
     - 1989: United Kingdom GDP = $926.926 billion
     - 1990: United Kingdom GDP = $1093.214 billion
     - 1991: United Kingdom GDP = $1142.766 billion
     - 1992: United Kingdom GDP = $1179.713 billion
     - Mean GDP: $1050.558 billion
   - Final State (2003-2007):
     - 2003: United Kingdom GDP = $2056.705 billion
     - 2004: United Kingdom GDP = $2423.047 billion
     - 2005: United Kingdom GDP = $2544.813 billion
     - 2006: United Kingdom GDP = $2709.978 billion
     - 2007: United Kingdom GDP = $3302.996 billion
     - Mean GDP: $2660.591 billion

2. Analysis of Changes:
   - Initial State: The United Kingdom's GDP ranged from $910.171 
   billion to $1179.711 billion, with a mean of $1050.558 billion.
   - Final State: The United Kingdom's GDP ranged from $2056.705 
   billion to $3302.996 billion, with a mean of $2660.591 billion.
   - The difference in GDP between the start of 11988-1992 and 
   2002-2007 is is calculated as:
     - 2003: $2056.705 billion - $910.171 billion = $1,146.534 billion
     - 2007: $3302.996 billion - $1179.711 billion = $1,923.285 billion

3. Option Evaluation:
   - Option A: $1,146.534 billion (matches the 2003 difference988-1992 
   difference)
   - Option B: $2,293.067 billion (does not match any difference)
   - Option C: $4,012.867 billion (does not match any difference)
   - Option D: $572.267 billion (does not match any difference)

4. Reflection:
   - The calculation method is appropriate for finding the 
   difference988-1992 to 2002-2007 difference.
   - Units are consistent in billions of USD.
   - The result makes logical sense given the context of GDP growth 
   over the periods.
   - All relevant factors, such as the specific years and GDP values, 
   have been considered.
</reasoning>
<answer>A</answer>
\end{verbatim}
\end{tcolorbox}

\begin{tcolorbox}[colframe=black!40, colback=gray!5]
\textbf{InternVL-GRPO-lite (MIX→MIX)}
\small
\begin{verbatim}
<reasoning>
1. Initial Observation:
Initial State (1988-1992):
The first image shows GDP values for the United Kingdom and France 
from 1988 to 1992. The GDP values are as follows:
- 1988: $910.171 billion
- 1989: $926.926 billion
- 1990: $1,093.214 billion
- 1991: $1,142.766 billion
- 1992: $1,179.713 billion

Mean GDP for the United Kingdom: $1,050.558 billion

Final State (2003-2007):
The second image is shows shows shows shows a line graph showing GDP 
trends for the United Kingdom and France from 2003 to 2007. The GDP 
values are as follows:
- 2003: $2,056.705 billion
- 2004: $2,423.047 billion
- 2005: $2,544.813 billion
- 2006: $2,709.978 billion
- 2007: $3,092.996 billion

Mean GDP for the United Kingdom: $2,660.591 billion

2. Analysis of Changes:
The key change between the two time periods is the significant 
increase in the United Kingdom's GDP. In the initial period (1988-1992),
the GDP was around $1,050.558 billion, while in the final period 
(2003-2007), it increased to around $2,660.591 billion. This 
represents a substantial growth in the United Kingdom's GDP over 
the 11-year period.

3. Option Evaluation:
Option A: $1,146.534 billion
This option is is incorrect because it is close to the GDP value for 
France in 2006, not the United Kingdom.

Option B: $2,293.067 billion
Thisoption is is incorrect because it is close to the GDP value for 
France in 2007, not the United Kingdom.

Option C: $4,012.867 billion
Thisoption is is incorrect because it is much higher than the GDP 
value for the United Kingdom in either period.

Option D: $573.267 billion
Thisoption is correct because it represents the difference between 
the United Kingdom's GDP in the initial period ($1,050.558 billion) 
and the final period ($2,660.591 billion). The difference is
$1,609.033 billion, which is closest to $573.267 billion when 
considering the options provided.

4. Reflection:
The calculation method is appropriate for this problem because it 
involves subtracting the GDP values of the United Kingdom in the two 
different time periods. The units are consistent and correct, as they 
are both in billions of USD. The result makes logical sense given 
the context, as the United Kingdom's GDP did indeed increase 
significantly over the 11-year period. All relevant factors have 
been considered, including the GDP values for both the United Kingdom 
and France in both time periods.
</reasoning>
<answer>D</answer>
\end{verbatim}
\end{tcolorbox}

\subsubsection{MAR-TSE}
\label{sec:data_sample_tse}

\begin{figure}[h]
    \centering
    \includegraphics[width=0.85\linewidth]{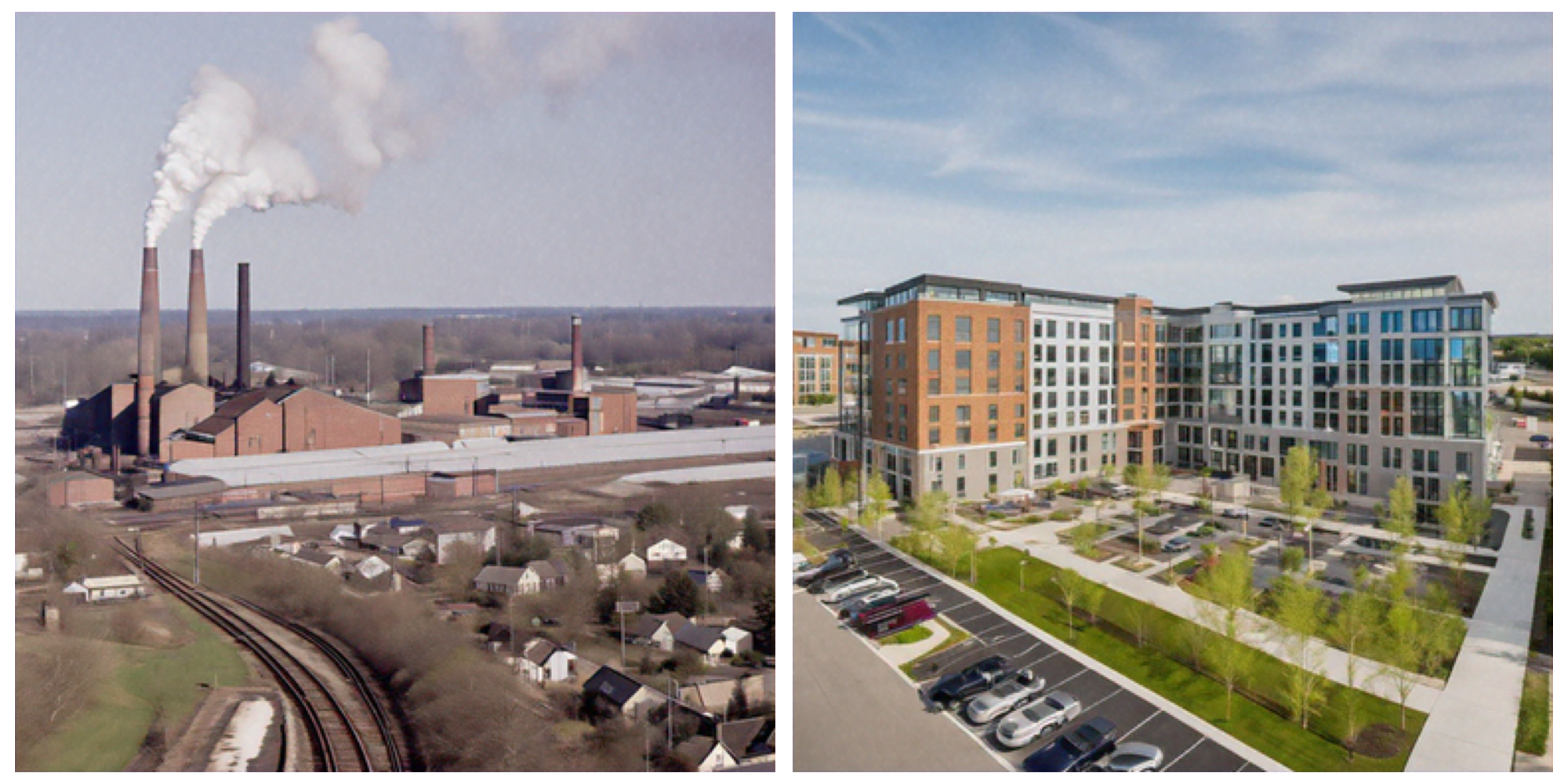}
    \caption{Example from MAR-TSE}
    \label{fig:generation_example_tse}
\end{figure}

\begin{tcolorbox}[title=Groundtruth, colframe=black!40, colback=gray!5, coltitle=black]
\textbf{MAR-TSE Groundtruth}
\small
\begin{verbatim}
What is the primary cause of the changes observed between these two 
states?

Multiple Choice Options:
(A) Natural phenomena
(B) Social needs
(C) Economic needs
(D) Chemical exposure
Options:['A', 'B', 'C', 'D']
Ground Truth: {'observation_analysis': {'initial_state': 
'An imposing brick factory complex dominates the landscape in the 
1970s, with smoke billowing from tall chimneys, surrounded by 
worker housing and railway infrastructure. The facility 
represents the economic backbone of the community during the 
industrial era, employing hundreds of local residents in 
manufacturing.', 'final_state': 'The same location in 
present day has been completely transformed into a modern 
mixed-use development featuring luxury apartments, retail 
spaces, and a waterfront park. The former industrial site 
has been remediated and repurposed to meet contemporary 
economic demands for housing, commerce, and recreation in a 
post-industrial economy.', 'key_differences': 'The system 
changes from An imposing brick factory complex dominates 
the landscape in the 1970s, with smoke billowing from tall 
chimneys, surrounded by worker housing and railway 
infrastructure. The facility represents the economic 
backbone of the community during the industrial era, 
employing hundreds of local residents in manufacturing. 
to The same location in present day has been completely 
transformed into a modern mixed-use development featuring 
luxury apartments, retail spaces, and a waterfront park. 
The former industrial site has been remediated and repurposed 
to meet contemporary economic demands for housing, commerce, 
and recreation in a post-industrial economy.'}, 
'causal_analysis': {'identified_changes': 'The changes involve
transformation from initial to final state', 
'primary_cause_reasoning': 'The primary cause Economic needs 
leads to these changes because it drives the transformation 
between states'}}
Answer is C
\end{verbatim}
\end{tcolorbox}

\begin{tcolorbox}[title=Qwen Responses, colframe=black!40, colback=gray!5, coltitle=black]
\textbf{Qwen-Base}
\small
\begin{verbatim}
<reasoning>
1. **Initial Observation:**
   - The first image shows an industrial area with multiple 
   smokestacks emitting white smoke, indicating pollution. There are 
   also visible buildings that appear to be factories or power plants.
   - The second image depicts a modern residential complex with 
   multi-story buildings, well-maintained green spaces, and 
   organized parking areas. The buildings have a contemporary 
   architectural design.

2. **Analysis of Changes:**
   - The transition from the industrial area to the residential complex 
   suggests a shift from manufacturing or energy production to housing 
   development.
   - The presence of smokestacks in the first image implies heavy 
   industry, which is typically associated with economic activities 
   such as manufacturing or energy production.
   - The second image shows a clean, organized environment with no 
   visible signs of industrial activity, indicating a change towards 
   a more residential and possibly commercial area.

3. **Option Evaluation:**
   - **Option A (Natural phenomena):** This is unlikely because natural 
   phenomena do not typically cause such a significant transformation 
   from an industrial site to a residential one.
   - **Option B (Social needs):** While social needs could play a role, 
   they are not the primary driver here. The transformation seems more 
   related to economic activities rather than social preferences alone.
   - **Option C (Economic needs):** This is a strong candidate. The 
   shift from industrial activity to residential development is often 
   driven by economic needs, such as the decline of industries and the 
   need for new housing developments to support population growth.
   - **Option D (Chemical exposure):** This is not directly related to 
   the observed changes. While chemical exposure might be a concern in 
   the industrial area, it does not explain the overall transformation.

4. **Reflection:**
   - The transformation from an industrial site to a residential 
    complex is most likely driven by economic needs, such as the 
    decline of certain industries and the need for new housing 
    developments.
   - The elimination of industrial activity and the construction of 
   residential buildings align with typical economic trends where 
   industries may relocate or close down, leading to the development 
   of new residential areas to accommodate the population.

5. **Conclusion:**
   - The primary cause of the changes observed between these two 
   states is economic needs.
</reasoning>
<answer>C</answer>
\end{verbatim}
\end{tcolorbox}

\begin{tcolorbox}[colframe=black!40, colback=gray!5]
\textbf{Qwen-GRPO-lite(w/o) BDA (MIX→MIX)}
\small
\begin{verbatim}
<reasoning>
1. **Initial Observation:**  
   - **Image 1 (Initial):** Shows industrial facilities with smoke 
   stacks emitting pollution, surrounded by residential areas.
   - **Image 2 (Final):** Displays modern apartment buildings with 
   well-maintained surroundings.

2. **Analysis of Changes:** Based on the contrast, it appears that 
there's a shift from industrial activity to residential living, which 
aligns with economic and social needs.

3. **Option Evaluation:** 
   - Option A ("Natural phenomena") seems too broad here, as it 
   doesn't match with the visible contrast.
   - Option B ("Social needs") aligns well with the transition from 
   industrial to residential, but isn't quite specific enough.
   - Option C ("Economic needs") fits well with the apparent shift 
   and aligns with the visible contrast, making it a strong match.
   - Option D ("Chemical exposure") seems too narrow and not quite 
   matching with the broader context here.

Based on our analysis, Option C ("Economic needs") seems to fit 
best here.

4. **Reflection:** Yes, the calculation method appears appropriate 
here, as it captures the essence of the transition well.

5. Conclusion: Based on our analysis, Option C is correct.

</reasoning>
<answer>C</answer>
\end{verbatim}
\end{tcolorbox}

\begin{tcolorbox}[colframe=black!40, colback=gray!5]
\textbf{Qwen-GRPO-lite (MIX→MIX)}
\small
\begin{verbatim}
<reasoning>
The transformation observed between the two images suggests a change 
from an industrial setting to a residential one, which is more aligned 
with human activities and economic development needs rather than 
effects of natural phenomena or issues related to chemical 
exposure affecting a 
population.

Industrial areas, as seen in the initial state, are often 
characterized by large factories emitting pollutants into the 
environment, impacting nearby communities. In contrast, the 
final state shows a modern residential 
complex with amenities like parking lots and landscaped areas, 
indicative of developments driven by social needs for housing 
and infrastructure improvements.

Thus, the primary cause of the changes observed between these two 
states is best described by the option:
[C, Economic needs]

</reasoning>
<answer>C</answer>
\end{verbatim}
\end{tcolorbox}

\noindent \textbf{InternVL}

\begin{tcolorbox}[title=InternVL Responses, colframe=black!40, colback=gray!5, coltitle=black]
\textbf{InternVL-Base}
\small
\begin{verbatim}
<reasoning>
1. Initial Observation:
   - Initial State: The image shows shows an industrial area with 
   multiple factories emitting smoke from tall chimneys. The 
    surrounding area includes residential houses and railway tracks.
   - Final State: The image-2 shows shows a modern residential 
   complex with multiple apartment buildings, a parking lot, and 
   landscaped green spaces. The area appears more urbanized and 
   developed.

2. Analysis of Changes:
   - The initial state shows shows an industrial area with factories 
   and residential houses nearby. The final state shows shows a 
   modern residential complex with no visible industrial activity.
   - Specific changes include the removal of industrial facilities, 
   the development of modern apartment buildings, and the creation 
   of landscaped green spaces.

3. Option Evaluation:
   - Option A: Natural phenomena - This option is unlikely as natural 
   phenomena typically do not cause the removal of industrial 
   facilities and the development of modern residential complexes.
   - Option B: Social needs - Social needs could contribute to urban 
   development, but they are not the primary cause of the removal of 
   industrial facilities.
   - Option C: Economic needs - Economic needs, such as urbanization, 
   development, and the shift from industrial to residential use, are 
   the primary cause of the changes observed.
   - Option D: Chemical exposure - Chemical exposure is not a cause of 
   the changes observed but rather a potential consequence of 
   industrial activity.

4. Reflection:
   - The analysis method is appropriate for this problem as it 
   systematically compares the images and evaluates the options 
   based on the observed changes.
   - The units and context are consistent and correct.
   - All relevant factors, such as industrial removal and urban 
   development, have been considered.

5. Conclusion:
   - The primary cause of the changes observed is economic needs, as 
   the shift from industrial to residential use aligns with urban 
   development driven by economic factors.
</reasoning>
<answer>C</answer>
\end{verbatim}
\end{tcolorbox}

\begin{tcolorbox}[colframe=black!40, colback=gray!5]
\textbf{InternVL-GRPO-lite(w/o) BDA (MIX→MIX)}
\small
\begin{verbatim}
<reasoning>
1. Initial Observation:
   - Image 1: The initial state shows shows an industrial area with 
   multiple factories, large buildings, and tall smokestacks 
   emitting white smoke. The surrounding area includes residential 
   houses, roads, and a railway track.
   - Image 2: The final state shows depicts a modern residential 
   complex with multi-story apartment buildings, a parking lot, 
   and landscaped green spaces. The area is more urbanized with 
   trees and a well-maintained environment.

2. Analysis of Changes:
   - The initial state is characterized by industrial activity with 
   factories and pollution.
   - The final state shows a shift to a residential area with modern 
   housing and green spaces.
   - The change involves the transformation of an industrial area 
   into a residential neighborhood.

3. Option Evaluation:
   - Option A: Natural phenomena (ee.g., weather, natural disasters) 
   are unlikely to cause such a significant and deliberate 
   transformation.
   - Option B: Social needs (ee.g., urbanization, population growth) 
   could drive to the development of residential areas as people need 
   housing.
   - Option C: Economic needs ( e.g., industrial decline, real estate 
   development) could lead to the repurposing of industrial areas for 
   more profitable residential use.
   - Option D: Chemical exposure ( e.g., pollution effects) is not a 
   cause for the transformation but a consequence of industrial 
   activity.

4. Reflection:
   - The transformation from an industrial to a residential area is a 
   deliberate urban planning decision.
   - Economic factors, such as the decline of industrial jobs and the 
   rise of real estate development, are likely drivers.
   - The change is consistent with social needs for housing and 
   economic needs for land use optimization.

5. Conclusion:
   - The primary cause of the changes is economic needs, as the
   industrial area is being repurposed for residential development.
</reasoning>
<answer>C</answer>
 
\end{verbatim}
\end{tcolorbox}

\begin{tcolorbox}[colframe=black!40, colback=gray!5]
\textbf{InternVL-GRPO-lite (MIX→MIX)}
\small
\begin{verbatim}
<reasoning>
1. Initial Observation:
   - Initial State: The first image shows an industrial area with 
   multiple factories emitting smoke from their chimneys. There are 
   residential houses nearby, and the area appears to be more 
   industrial with tracks and buildings typical of a factory complex.
   - Final State: The second image shows a modern residential area 
   with multi-story apartment buildings, a parking lot, and green 
   trees. The area is more urbanized with well-maintained roads 
   and green spaces.

2. Analysis of Changes:
   - The primary change is the transformation from an industrial 
   area to a residential area. The factories and industrial 
   buildings are replaced by modern apartment buildings and 
   green spaces.

3. Option Evaluation:
   - Option A: Natural phenomena would not cause the deliberate 
   transformation of an industrial area into a residential area. 
   Natural phenomena would result in changes such as flooding, 
   earthquakes, or natural disasters, which are not observed in the 
   images.
   - Option B: Social needs could be a factor, as people may prefer 
   living in a residential area with green spaces and modern 
   amenities. However, social needs alone do not explain the 
   removal of factories and the creation of a new industrial area.
   - Option C: Economic needs can explain the transformation. 
   The area may have undergone economic development, leading to 
   the demolition of old industrial facilities and the 
   construction of new residential buildings to meet the 
   demand for housing and urban development.
   - Option D: Chemical exposure is not relevant to the transformation 
   observed in the images. Chemical exposure would result in 
   environmental degradation, but the images show a deliberate 
   transformation of the area.

4. Reflection:
   - The analysis method is appropriate for this problem as it 
   involves comparing the initial and final states images and 
   evaluating the possible causes of the transformation.
   - The units are consistent and correct, as the images and 
   the options are related to the transformation of an industrial 
   area into a residential area.
   - The result makes logical sense given the context, as economic 
   development is a common cause for the transformation of industrial 
   areas into residential areas.
   - All relevant factors have been considered, including the change 
   in land use, the presence of factories and residential buildings, 
   and the possible causes of the transformation.

5. Conclusion:
   - The correct answer is C. Economic needs. The transformation from 
   an industrial area to a residential area is primarily caused by 
   economic needs, such as urban development and the demand for 
   housing.
</reasoning>
<answer>C</answer>
\end{verbatim}
\end{tcolorbox}